\documentclass[11pt]{article}

\newcommand{\hy}{\hat{y}}
\newcommand{\err}{{\bf er}}
\newcommand{\N}{{\bf N}}
\newcommand{\R}{{\bf R}}
\newcommand{\sam}{\mbox{\rm sam}}
\newcommand{\VCdim}{\mathrm{VCdim}}

\newcommand{\fat}{\mathrm{fat}}
\newcommand{\fatV}{\mathrm{fatV}}
\newcommand{\sfat}{\mathrm{sfat}}
\newcommand{\fun}[3]{ #1 : #2 \to #3}

\newcommand{\floor}[1]{{\left\lfloor #1 \right\rfloor}}
\newcommand{\ceil}[1]{{\left\lceil #1 \right\rceil}}
\newcommand{\rest}[2]{{ #1 }_{|_{ #2 }}}
\newcommand{\qed}{\hfill\protect\raisebox{.3ex}{\framebox[0.5em]{\rule{0em}{.6ex}}}}
\newcommand{\comb}[2]{{{ #1 } \choose { #2 }}}
\newcommand{\first}{{\mbox{\rm\tiny first}}}
\newcommand{\last}{{\mbox{\rm\tiny last}}}

\usepackage{graphicx} 
\usepackage{fullpage}
\usepackage{amsmath}
\usepackage{amsfonts}
\usepackage{xcolor}
\usepackage{bbm}

\newenvironment{proof}{{\bf Proof:}}{\qed}

\newcommand{\indicator}{\mathbbm{1}}

\newcommand*{\eqdef}{\stackrel{\textup{def}}{=}}

\newcommand{\cL}{{\cal L}}

\newcommand{\tA}{\tilde{A}}

\newcommand{\E}{{\bf E}}

\renewcommand{\Pr}{{\bf Pr}}

\newtheorem{theorem}{Theorem}
\newtheorem{proposition}[theorem]{Proposition}
\newtheorem{lemma}[theorem]{Lemma}
\newtheorem{corollary}[theorem]{Corollary}

\usepackage{titling}
\usepackage[sectionbib]{chapterbib}
\newcommand{\hspeq}{\hspace{5pt}}

\begin{document}
\title{\bf Prediction, Learning, Uniform Convergence,\\
and Scale-sensitive Dimensions}

\author{
Peter L. Bartlett\thanks{Also affiliated with University of California, Berkeley.}\hspace{5pt}
and
Philip M. Long  \\
Google \\
% $\{$peterbartlett,plong$\}$@google.com 
}
\date{}

\maketitle

\section*{\centering Abstract \\}
We present a new general-purpose algorithm for learning classes of
$[0,1]$-valued functions in a generalization of the prediction model, and
prove a general upper bound on the expected absolute error of this
algorithm in terms of a scale-sensitive generalization of the Vapnik
dimension proposed by Alon, Ben-David, Cesa-Bianchi and Haussler.  We give
lower bounds implying that our upper bounds cannot be improved by more than
a constant factor in general.  We apply this result, together with
techniques due to Haussler and to Benedek and Itai, to obtain new upper
bounds on packing numbers in terms of this scale-sensitive notion of
dimension.  Using a different technique, we obtain new bounds on packing
numbers in terms of Kearns and Schapire's fat-shattering function.  We show
how to apply both packing bounds to obtain improved general bounds on the
sample complexity of agnostic learning.  For each $\epsilon > 0$, we
establish weaker sufficient and stronger necessary conditions for a class
of $[0,1]$-valued functions to be agnostically learnable to within
$\epsilon$, and to be an $\epsilon$-uniform Glivenko-Cantelli class.

This is a manuscript that was accepted by JCSS, together with a correction.

\emptythanks
\thispagestyle{empty}

\newpage
\setcounter{page}{1}

\title{\bf Prediction, Learning, Uniform Convergence,\\
and Scale-sensitive Dimensions \\
(correction) \\}

\author{
Peter L. Bartlett\thanks{Also affiliated with University of California, Berkeley.}\hspace{5pt}
and
Philip M. Long  \\
Google \\
% $\{$peterbartlett,plong$\}$@google.com 
}
\date{}

\maketitle

\begin{abstract}
We describe a flaw in the proof of Theorem 1 of \cite{bartlett1998prediction} pointed out by Ishaq Aden-Ali, Yeshwanth Cherapanamjeri, Abhishek Shetty, and Nikita Zhivotovski, and provide a 
new proof with an improved constant.
\end{abstract}

\section{Introduction and definitions}

Recently, Ishaq Aden-Ali, Yeshwanth Cherapanamjeri, Abhishek Shetty, and Nikita Zhivotovski alerted us to the
fact that the proof of Theorem 1 of \cite{bartlett1998prediction}
is flawed.  In this note, we describe the mistake, and provide a revised proof
analyzing a similar algorithm, with a better constant.  
A related result, with a worse constant, was obtained by 
Aden-Ali et al \cite{ACSZ23}.

Here are the relevant definitions from
\cite{bartlett1998prediction}.  For a domain $X$,
a prediction
strategy is a mapping from 
$(\cup_m (X \times [0,1])^m) \times X$ to $[0,1]$.
For a set $F$ of functions from $X$ to $[0,1]$,
\[
\cL(F,m) = \inf_A \sup_{D,f} 
    \E_{x_1,...,x_m \sim D^m} \big[\big| A(x_1, f(x_1),...,(x_{m-1}, f(x_{m-1})),x_m)
                  - f(x_m)
           \big|\big],
\]
where the infimum ranges over prediction strategies, 
and the supremum ranges over probability distributions
$D$, and $f \in F$.  

A concept-with-don't-cares (CWDC) strategy is a mapping from 
$(\cup_m (X \times \{ 0, \star, 1\})^m) \times X$ to $\{ 0,1 \}$.

For a function $f$ from $X$ to $[0,1]$, a scale parameter
$\gamma > 0$, and a threshold $r$,
$\psi_{r,\gamma,f} (x) = 
\psi_{r,\gamma} (f(x)),
$
where, for all $y \in [0,1]$, we define
\[
\psi_{r,\gamma}(y) =
  \left\{
    \begin{array}{ll}
    1 & \mbox{ if $y \geq r + \gamma$} \\
    \star & \mbox{ if $r - \gamma < y < r + \gamma$} \\
    0 & \mbox{ if $y \leq r - \gamma$.} \\
    \end{array}
  \right.
\]

For $\gamma > 0$,
the scale-sensitive definition $\fatV_{\gamma}(F)$
is the largest $d$ for which there
an $r \in [0,1]$ and 
$x_1,...,x_d \in X$ such that
\[
\{ 0,1 \}^d \subseteq 
  \{ (\psi_{r,\gamma}(f(x_1)),...,\psi_{r,\gamma}(f(x_d)))
       :
      f \in F
      \}.
\]

\section{The issue}

The Algorithm $A$ analyzed in the
flawed proof of Theorem 1 of \cite{bartlett1998prediction}
learns a class $F$ of $[0,1]$-valued functions
by using subroutine CWDC strategies $B_r$ that learn 
$\psi_{r,\gamma}(F) = \{ \psi_{r,\gamma,f}: f \in F \}$.

The proof of Lemma 4 in \cite{bartlett1998prediction} shows that, for all $r$, there exists a CWDC strategy $B_r$ such that for all
$D$, and for all $f \in F$, we have
\begin{align}
\nonumber
& \Pr_{x_1,...,x_m \sim D^m}
\big[ 
B_r\big(
   ((x_1, \psi_{r,\gamma,f}(x_1)),
   ...,
   ((x_{m-1}, \psi_{r,\gamma,f}(x_{m-1})),
   x_m
    \big)
    \neq \psi_{r,\gamma,f}(x_m) \\
    \nonumber
& \hspace{2in}
    \mbox{ and }
    \psi_{r,\gamma,f}(x_m) \neq \star
\big] \\
\label{e:fixed_r}
 & \hspace{0.5in} \leq 
 \frac{\fatV_{\gamma}(F)}{m}.
\end{align}

The algorithm $A$ analyzed in Theorem 1 of \cite{bartlett1998prediction} uses the subroutine algorithms
$B_{1/2}, B_{1/4}, B_{3/4},...$ as noisy oracles to perform binary search, using
\eqref{e:fixed_r} to control the noise.  Its proof included a claim that 
\eqref{e:fixed_r} 
implies
%bounded an event $E_j$ above by 
that the probability
that the $j$th comparison of the binary search is incorrect, and not within $\gamma$ of
the boundary, is at most
$\fatV_{\gamma}(F)/m$.  
This is not valid because \eqref{e:fixed_r} provides
a bound that holds when $r$ is fixed before sampling $x_1,...,x_m$, but
the threshold used in the $j$th round of the binary search of
Algorithm $A$ depends on the random choice of $x_m$.

\section{A new analysis of a new algorithm}

In this section, we prove the following,
which improves on
the $2 \fatV_F(\gamma)/m + \gamma$
bound claimed in
Theorem 1
of \cite{bartlett1998prediction}.
\begin{theorem}
\label{t:fix}
For any set $F$ of functions from 
$X$ to $[0,1]$, for any $\gamma > 0$, and any
positive integer $m$, we have
\[
\cL(F,m) \leq \frac{\fatV_F(\gamma)}{m} + \gamma.
\]
\end{theorem}
\begin{proof}
For an arbitrarily small $\tau > 0$, we will define an Algorithm $\tA_{\tau}$,
and prove that its expected error is less than
$\fatV_F(\gamma)/m + \gamma + 2 \tau$.
Fix $\tau > 0$ and $m$.

% Recall that, for $s > 0$, $Q_{\tau}(s) \eqdef \tau \lfloor s/\tau \rfloor$
% and $Q_{\tau}(S) \eqdef \{ Q_{\tau}(s) : s \in S \}$.

Algorithm $\tA_{\tau}$ calls $B_r$ for all 
$r \in R \eqdef \{ \tau, 2 \tau, \ldots, \lfloor 1/\tau \rfloor \tau  \}$.
Denote the various predictions made by these subroutine
algorithms by $\{ b_r : r \in R \}$.
The prediction of $\tA_{\tau}$ is then 
\[
\hy = \tau \sum_{r \in R} b_r = \tau | \{ r \in R : b_r = 1 \} |.
\]
(To motivate this, it may be helpful to think 
of $b_r$ as a guess at whether $y \geq r$;
if all of these guesses are correct, then
$b_r = 1 \Leftrightarrow y \geq r$, and the proportion
of such values of $r$ is, up to rounding
error, $y$.)

We claim that, for any $y \in [0,1]$ and
for any predictions $\{ b_r : r \in R \}$,
\begin{align}
\label{e:by_binary}
| y - \hy | 
   & < 2 \tau + \gamma + \tau | \{ r \in R :b_r \neq \psi_{r,\gamma}(y) \mbox{ and } \psi_{r,\gamma}(y) \neq \star \} |.
\end{align}

To see this, first note that
$\hy$ is an integer multiple of
$\tau$, so
\begin{align*}
\nonumber
| \hy - y |
 & \leq \tau + | \hy - \tau \lfloor y/\tau \rfloor | \\
\nonumber
 & = \tau + \tau \left| \sum_{r \in R} (b_r - \indicator_{y \geq r}) \right| \\
    \nonumber
% \label{e:inside_outside}
 & \leq \tau 
 + \tau \left|\sum_{r \in R: |y - r| < \gamma} (b_r - \indicator_{y \geq r})
    \right| 
 +\tau \left| \sum_{r \in R: |y - r| \geq \gamma} (b_r - \indicator_{y \geq r})
    \right|. \\
\end{align*}
Since $\left|\sum_{r \in R: |y - r| < \gamma} (b_r - \indicator_{y \geq r})
    \right|$ is maximized when
$b_r = 0$ for all $r$ in this sum, 
where it takes the value $\left|\left\{r\in R: y-\gamma<r\le y\right\}\right| \leq \lceil\gamma/\tau\rceil < 1 + \gamma/\tau$,
we have
\begin{align*}
\nonumber
| \hy - y |
    \nonumber
% \label{e:inside_outside}
 & < 2 \tau + \gamma
 +\tau \left| \sum_{r \in R: |y - r| \geq \gamma} (b_r - \indicator_{y \geq r})
    \right| \\
    \nonumber
% \label{e:inside_outside}
 & \leq 2 \tau + \gamma
 +\tau \sum_{r \in R: |y - r| \geq \gamma} \left| b_r - \indicator_{y \geq r}
    \right| \\
% \label{e:inside_outside}
 & = 2 \tau 
 + \gamma
 +\tau | \{ r \in R :b_r \neq \psi_{r,\gamma}(y) \mbox{ and } \psi_{r,\gamma}(y) \neq \star \} |,
\end{align*}
establishing \eqref{e:by_binary}.

For each $r$ (note that, here, $r$ is not random), 
\eqref{e:fixed_r} implies
\[
\Pr[b_r \neq \psi_{r,\gamma}(f(x_m)) \mbox{ and } \psi_{r,\gamma}(f(x_m)) \neq \star]
 \leq \frac{\fatV_F(\gamma)}{m}.
\]

Thus,
\begin{align*}
\E[
| \{ r \in R : 
  b_r \neq \psi_{r,\gamma}(f(x_m)) \mbox{ and } \psi_{r,\gamma}(f(x_m)) \neq \star
   \} | ] &
 \leq |R| \times \frac{\fatV_F(\gamma)}{m}  
 \leq \frac{\fatV_F(\gamma)}{\tau m}.  \\
\end{align*}

Combining this with \eqref{e:by_binary} completes the proof.
\end{proof}

The original uncorrected paper is attached below.

\bibliographystyle{plain}
\bibliography{bib}

\clearpage
\emptythanks

\setcounter{page}{1}
\setcounter{section}{1}
\setcounter{theorem}{0}

% 0preamble.tex
% title, abstract

\title{\bf Prediction, Learning, Uniform Convergence,\\
and Scale-sensitive Dimensions \\
(original accepted manuscript) \\}

\author{
Peter L. Bartlett \\
Australian National University \\
\and
Philip M. Long  \\
National University of Singapore \\
% $\{$peterbartlett,plong$\}$@google.com 
}
\date{March 24, 1997}

\maketitle
\setcounter{page}{0}
\thispagestyle{empty}

\section*{\centering Abstract \\}
We present a new general-purpose algorithm for learning classes of
$[0,1]$-valued functions in a generalization of the prediction model, and
prove a general upper bound on the expected absolute error of this
algorithm in terms of a scale-sensitive generalization of the Vapnik
dimension proposed by Alon, Ben-David, Cesa-Bianchi and Haussler.  We give
lower bounds implying that our upper bounds cannot be improved by more than
a constant factor in general.  We apply this result, together with
techniques due to Haussler and to Benedek and Itai, to obtain new upper
bounds on packing numbers in terms of this scale-sensitive notion of
dimension.  Using a different technique, we obtain new bounds on packing
numbers in terms of Kearns and Schapire's fat-shattering function.  We show
how to apply both packing bounds to obtain improved general bounds on the
sample complexity of agnostic learning.  For each $\epsilon > 0$, we
establish weaker sufficient and stronger necessary conditions for a class
of $[0,1]$-valued functions to be agnostically learnable to within
$\epsilon$, and to be an $\epsilon$-uniform Glivenko-Cantelli class.

\vfill
\newpage

% 1intro.tex

\section{Introduction}

In the prediction model studied in this paper, a $[0,1]$-valued function
$f$ chosen from some known class $F$ is hidden from the learner, the
learner is given examples of $f$ evaluated at $m-1$ elements of the domain
of $f$ that were chosen independently at random according to an arbitrary,
unknown distribution, another random point $x$ is chosen, and the learner
is required to output a prediction $\hy$ of $f(x)$.  The learner is
penalized by $|\hy-f(x)|$.  This can be viewed as a model of on-line
learning, and is the straightforward generalization of the prediction model
of Haussler, Littlestone and Warmuth \cite{HLW} to real-valued functions.

In this paper, we begin by introducing a new general-purpose prediction
strategy that uses a binary search to divide the problem of real-valued
prediction into a number of binary-valued prediction problems.  We give
bounds on the expected error of this strategy in terms of
% $\fatV_F$, the
$\fatV$, the
scale-sensitive generalization of the Vapnik dimension introduced by Alon,
Ben-David, Cesa-Bianchi and Haussler~\cite{abch} (which is similar to a
notion introduced by Kearns and Schapire \cite{p-concepts}), and show that
no algorithm can improve on these bounds in general by more than a constant
factor.

A packing number for a class of functions measures, in a certain sense,
the largest number of significantly different behaviors functions in the
class can have on a set of points of a given size.
We apply the above prediction bound, together with ideas due to Haussler
\cite{haussler-sphere} and Benedek and Itai \cite{benedek}, to obtain new
bounds on packing numbers in terms of $\fatV$.  

In agnostic learning \cite{haussler-decision,kss}, a distribution on $X
\times [0,1]$ is unknown, and the learner, given examples drawn according
to this distribution, tries to find a function $h$ from $X$ to $[0,1]$ so
that, with probability at least $1-\delta$, the expectation of $|h(x)-y|$
is at most $\epsilon$ larger than the minimum of this expectation over
functions in some touchstone class $F$.
We combine our new packing bound with the techniques
of another paper of Haussler~\cite{haussler-decision} 
to prove an upper bound\footnote{The $5$ in this
bound can be replaced with any constant greater than $4$.}
of
% $
\[
O\left( \frac{1}{\epsilon^2} \left(\frac{\fatV_F(\epsilon/5)}{\epsilon} \log \frac{1}{\epsilon}
                              +\log\frac{1}{\delta}\right)\right)
\]
% $
on the sample complexity of $(\epsilon,\delta)$-agnostic learning 
$F$.
This improves on the bound
%\footnote{The constant $384$ in that
%result can be brought arbitrarily close to $8$ if certain calculations
%are done carefully, for example as we do in this paper,
%but the remaining factor of two improvement in the scale at which
%the finiteness of $\fatV$ is sufficient for learning is due to our
%improved packing bounds.}
of
% $
\[
O\left( \frac{1}{\epsilon^2} \left(\frac{\fatV_F(\epsilon/384)}{\epsilon}\log^2 \frac{\fatV_F(\epsilon/384)}{\epsilon}
                              +\log\frac{1}{\delta}\right)\right)
\]
% $
that is a straightforward consequence of the results
of~\cite{abch} (see \cite{blw}).

Next, using a different technique, we obtain a new packing bound 
in terms of Kearns and Schapire's fat-shattering function.  
This leads to a bound\footnote{This $5$ can also be replaced with
any constant greater than $4$.}
of
% $
\[
O\left( \frac{1}{\epsilon^2} \left(\fat_F(\epsilon/5) \log^2 \frac{1}{\epsilon}
                              +\log\frac{1}{\delta}\right)\right)
\]
% $
on the sample complexity of $(\epsilon,\delta)$-agnostic learning $F$.
This improves on the dependence on $\fat_F$
of the bound
%\footnote{The $192$ here can be brought
%arbitrarily close to $4$ with care.}
% $
\[
O\left( \frac{1}{\epsilon^2} 
\left(\fat_F(\epsilon/192) \log^2 \frac{\fat_F(\epsilon/192)}{\epsilon}
                              +\log\frac{1}{\delta}\right)\right)
\]
% $
that follows from the packing bound of \cite{abch} (see \cite{blw}).

In previously derived bounds on the sample complexity of agnostic learning
in terms of scale-sensitive notions of dimension, the scale at which the
dimension was measured was a large constant factor finer than the relative
accuracy to which the learner was learning.
In this paper, we
% PLB 6/10/95: we investigate, and begin to answer.
% begin to
investigate the question of at what scale the dimension needs to
be finite for a class of functions to be agnostically learnable to within
relative accuracy $\epsilon$ (also to be an $\epsilon$-uniform
Glivenko-Cantelli class).  Our results narrow the range between necessary
and sufficient ``scales'' to a factor of $2$.  Our weaker sufficient
conditions are proved using a new general-purpose prediction strategy
that directly makes use of a cover of the function class.
For $\epsilon$-agnostic learning with respect to a class $F$,
this strategy takes a sequence of labelled examples and a single
unlabelled example,
and constructs an $(\epsilon-\alpha)$-cover of the restriction of the
function class $F$ to the examples.
(Here $\alpha$ can be made as close to zero as desired.)
Then the strategy divides the sample into two parts,
and selects the function in the cover that has minimal error on the
first half of the sample.
We show that if this function is used to predict the labels of the
examples in the last half of the sample,
the expected error is within (approximately) $\epsilon$ of the minimal
error.
A standard technique converts this to an $\epsilon$-agnostic learning
algorithm.

% 2def.tex

\section{Definitions}
\label{s:defs}

\subsection{Definitions for the prediction model}
For a set $X$, a {\em prediction strategy} is a mapping from 
$(\cup_m (X \times [0,1])^m) \times X$ to $[0,1]$.  
Let ${\cal P}_X$ be the set of all prediction strategies, and
let ${\cal D}_X$ be the set of all probability distributions
on $X$.
For each set $F$ of functions from $X$ to $[0,1]$, and
each positive integer $m$, define\footnote{Throughout,
we ignore issues of measurability.
The reader may assume that $X$ is countable,
but significantly weaker assumptions,
like those of Pollard's~\cite{pollard} Appendix C,
suffice.}
${\cal L}(F,m)$ as
% the infimum over $A$ in ${\cal P}_X$ of the
% supremum over $D$ in ${\cal D}_X$ and $f$ in $F$ of the expectation of
% \[
% |A(((x_1,f(x_1)),...,(x_{m-1},f(x_{m-1}))),x_m)-f(x_m)|,
% \]
% where $x_1,\ldots,x_m$ is chosen according to $D^m$.
\[
{\cal L}(F,m) =
        \inf_{A\in {\cal P}_X}
        \sup
        \int_{X^m} |A(((x_1,f(x_1)),...,(x_{m-1},f(x_{m-1}))),x_m)-f(x_m)|
        \,dD^m(x_1,\ldots,x_m),
\]
where the supremum is over all $D$ in ${\cal D}_X$ and $f$ in $F$.  That
is, ${\cal L}(F,m)$ is the worst-case expected error of the best prediction
strategy.  This is a generalization of the $\{ 0,1 \}$ prediction model
of~\cite{HLW} to $[0,1]$-valued functions.

\subsection{Definitions for the agnostic learning model}
Define a {\em learner} for a set $X$ to be
a mapping from $\cup_{n \in \N} (X \times [0,1])^n$ to
$[0,1]^X$, i.e.\ to take a sequence of labelled examples, and
output a hypothesis.
If $h$ is a $[0,1]$-valued function defined on $X$, and
$P$ is a probability distribution over $X \times [0,1]$,
define the {\em error of $h$ with respect to $P$} as
\[
\err_P(h) = \int_{X\times [0,1]} |h(x) - y|\, dP(x,y).
\]
Suppose $F$ is a class of $[0,1]$-valued functions defined on $X$,
$0<\epsilon,\delta<1$ and $m\in\N$.  We say a learner $A$ {\em
$(\epsilon,\delta)$-learns in the agnostic sense with respect to $F$ from
$m$ examples} if, for all distributions $P$ on $X\times [0,1]$,
\[
P^m\left\{w\in\left(X\times[0,1]\right)^m:
\err_P(A(w))\ge \inf_{f\in F}\err_P(f) + \epsilon\right\}<\delta.
\]
For $\epsilon > 0$, the function class $F$ is {\em $\epsilon$-agnostically
learnable} if there is  a function $m_0:(0,1)\to \N$
such that, for all $0<\delta<1$, 
there is a learner $A$ which $(\epsilon,\delta)$-learns
in the agnostic sense with respect to $F$ from $m_0(\delta)$
examples.  

\subsection{Definition of $\epsilon$-uniform GC-classes}
For $\epsilon, \delta > 0$, a set $X$ and a set $F$ of functions
from $X$ to $[0,1]$, if ${\cal D}_X$ is the set of all probability
distributions over $X$, define
\begin{eqnarray*}
m_{{\mathrm{GC}},F} (\epsilon,\delta)
        & = &
        \min \left\{\rule[-2mm]{0cm}{8mm}
        n : \forall m\geq n,\forall D\in{\cal D}_X,
\right.\\ && \hspace{1cm} \left.
                D^m  \left\{ (x_1,...,x_m)\!:\! \exists f \in F,
                \left| \frac{1}{m}\left( \sum_{i=1}^m f(x_i) \right)
                        \!-\! \int_X f(u)\;dD(u) \right| > \epsilon\right\}
                                      \leq \delta \right\}.
\end{eqnarray*}
If the minimum doesn't exist, then 
$m_{\mathrm{GC},F}(\epsilon,\delta) = \infty$.
If, for all $\delta > 0$, $m_{\mathrm{GC},F}(\epsilon,\delta)$ is finite, then
$F$ is said to be an $\epsilon$-uniform GC- (Glivenko-Cantelli) class.

\subsection{Packing and Covering}
For each $n \in \N$, define $\fun{\ell_1}{\R^n \times \R^n}{\R}$ by
\[
\ell_1 (v,w) = \frac{1}{n} \sum_{i=1}^n |v_i - w_i|.
\]
For $S \subseteq \R^n$, define ${\cal N}(\epsilon,S)$ to be the
size of the smallest set $T \subseteq \R^n$ such that
for all $v \in S$, there is a $w \in T$ such that
$\ell_1 (v,w) \leq \epsilon$.  
Call such a $T$ an {\em $\epsilon$-cover}
of $S$.  Define ${\cal M}(\epsilon,S)$ to be the
size of the largest subset $T$ of $S$ such that for any
two elements $v,w$ of $T$, $\ell_1(v,w) > \epsilon$.
We will make use of the following well-known bounds that hold
for all $n$, $S \subseteq \R^n$.
\begin{equation}
\label{e:wellknown}
{\cal M}(2 \epsilon, S) \leq {\cal N}(\epsilon, S) 
                        \leq {\cal M}(\epsilon,S).
\end{equation}

\subsection{Quantizing}
For $\alpha > 0$ and $u \in \R$,
let $Q_{\alpha}(u)$ denote the quantized version of $u$, with
quantization width $\alpha$. That is,
define $Q_{\alpha}(u) = \alpha \lfloor u/\alpha \rfloor$.  
Let $Q_{\alpha}([0,1]) = \{ Q_{\alpha}(u) : u \in [0,1] \}$.
For $v \in \R^n$, define 
$Q_{\alpha}(v) = (Q_{\alpha}(v_1),...,Q_{\alpha}(v_n))$, and
similarly,
for a function $f$ from some set $X$ to $\R$, define
$\fun{Q_{\alpha}(f)}{X}{\R}$ by
$(Q_{\alpha}(f))(x) = Q_{\alpha}(f(x))$.
Finally, for a set $F$ of such functions, define 
$Q_{\alpha}(F) = \{ Q_{\alpha}(f) : f \in F \}.$

\subsection{Definitions relating to \protect$\fat$}
For $m \in \N$, $S \subseteq [0,1]^m$, and $\gamma>0$, we say
$S$ $\gamma$-fatly shatters a sequence
$(i_1,r_1),...,(i_d,r_d)$
of elements of $\{1,\ldots,m\}\times[0,1]$ if for all
$(b_1,...,b_d) \in \{ 0,1 \}^d$
there is a $v \in S$ such that for all $j \in \{ 1,...,d\}$,
\[
\begin{array}{ll}
v_{i_j} \geq r_j + \gamma & \mbox{if $b_j = 1$} \\
v_{i_j} \leq r_j - \gamma & \mbox{if $b_j = 0$.} \\
\end{array}
\]
We then define $\fat_S(\gamma)$ to be the length of the longest
sequence $\gamma$-fatly shattered by $S$.  
For a set $F$ of functions from $X$ to $[0,1]$, and
a finite sequence $\xi = (x_1,...,x_n)$ of elements
of $X$, define the restriction of $F$ to $\xi$ to be
\[
\rest{F}{\xi} = \{ (f(x_1),...,f(x_n)) : f \in F \}.
\]
We define $\fat_F(\gamma)$ to be the maximum, over all finite sequences
$\xi$ of elements of $X$, of $\fat_{\rest{F}{\xi}}(\gamma)$.  (This
was called the fat-shattering function in~\cite{blw}, and was
defined by Kearns and Schapire \cite{p-concepts}.)

\subsection{Definitions relating to \protect$\fatV$}
For each $r\in[0,1]$ and $\epsilon>0$, 
define $\fun{\psi_{r,\epsilon}}{[0,1]}{\{ 0,\star,1\}}$
by 
\[
\psi_{r,\epsilon} (y) = \left\{ \begin{array}{ll}
                                     1 & \mbox{if $y \geq r + \epsilon$} \\
                                     \star & \mbox{if $|y-r| < \epsilon$} \\
                                     0 & \mbox{if $y \leq r - \epsilon$.}
                                \end{array} \right.
\]
For a function $f$ from $X$ to $[0,1]$, define 
$\fun{\psi_{r,\epsilon}(f)}{X}{\{ 0,\star,1\}}$ by
\[
(\psi_{r,\epsilon}(f))(x) = \psi_{r,\epsilon}(f(x)),
\]
and for a set $F$ of such functions, define 
\[
\psi_{r,\epsilon}(F) = \{ \psi_{r,\epsilon}(f) : f \in F \}.
\]
We say $x_1,\ldots,x_d$ in $X$ are
$\gamma$-fatly Vapnik-shattered by $F$ if
there is an $r\in[0,1]$ such that
\[
\{ 0,1 \}^d \subseteq 
   \{ (\psi_{r,\gamma}(f(x_1)),...,
   \psi_{r,\gamma}(f(x_d))) : f \in F \}.
\]
Define $\fatV_F (\gamma)$ to be the length of the
longest sequence $\gamma$-fatly Vapnik-shattered by $F$.
(This dimension was first studied in~\cite{abch}.)

Notice that $\fat_F(\gamma)$ and $\fatV_F(\gamma)$ are both
non-increasing functions of $\gamma$.

% 2predict.tex

\section{Prediction of $[0,1]$-valued functions and $\fatV$}
\label{s:binary}

This section describes our general-purpose prediction strategy
and shows that it is nearly optimal.
The first theorem of the paper gives the bound for
the worst-case expected error incurred by this strategy.

\begin{theorem}
\label{t:binfun}
Choose a set $F$ of functions from $X$ to $[0,1]$, 
$\gamma > 0$, and a positive integer $m$.  Then
\[
{\cal L}(F,m) \leq \frac{2 \fatV_F (\gamma)}{m} + \gamma.
\]
\end{theorem}

Fix a set $X$.  Theorem~\ref{t:binfun} is proved by considering an
algorithm that generates its prediction using binary search (details
are given below).  It uses subalgorithms to predict whether $f(x_m)$
is above or below $1/2$, above or below $1/4$ and $3/4$, and so on.
To analyze these subalgorithms, we would like to show that, for
example, the set of possible ``above-below 1/2'' behaviors is not very
rich.  But a bound on $\fatV_F (\gamma)$ only provides information
about the richness of behaviors at least $\gamma$-above and
$\gamma$-below $1/2$.  On the other hand, if $\gamma$ is small, the
binary search algorithm can tolerate incorrect guesses if the truth
is within $\gamma$ of $1/2$, so, in a sense, we don't care about the
correctness of predictions in such cases.

Therefore, we will consider a model of learning which might be called
concept-with-don't-care's learning.  Here, what is learned is a
function from $X$ to $\{ 0,\star,1 \}$.  The $\star$ is interpreted as
a ``don't care'' value, in that an incorrect prediction of the value
of the function does not count against the learning algorithm if that
value is $\star$.  Also, when we generalize the VC-dimension, a notion
of the richness of a class of $\{ 0,1 \}$-valued functions, loosely
speaking, the $\star$'s will not contribute toward a certain class
being considered rich.

Formally, define a concept-with-don't-care's (CWDC) strategy to be a
mapping from
\[
\left(\bigcup_m \left(X\times\{0,\star,1\}\right)^m\right)\times X
\]
 to
$\{0,1\}$.
Let ${\cal B}_X$ be the set of all CWDC strategies.
For each set $G$ of functions from $X$ to $\{ 0,\star,1 \}$,
% let
% $M(G,m)$ be the infimum over $B$ in ${\cal B}_X$ of the supremum over
% $D$ in ${\cal D}_X$ and $g$ in $G$ of the probability under $D^m$ of a
% sequence $(x_1,\ldots,x_m)$ for which
% $g(x_m) \neq \star$ and
% \[
% B(((x_1,g(x_1)),...,(x_{m-1},g(x_{m-1}))),x_m) \neq g(x_m).
% \]
define $M(G,m)$ as the worst case mistake probability of the
best CWDC prediction strategy in $G$,
\begin{eqnarray*}
M(G,m)
        & = &   \inf_{B \in {\cal B}_X} \sup
                D^m \left\{ (x_1,...,x_m) : g(x_m) \neq \star \right.\\
        &&              \hspace{1in} \left.\mbox{ and }
                B(((x_1,g(x_1)),...,(x_{m-1},g(x_{m-1}))),x_m) \neq g(x_m)
                        \right\},
\end{eqnarray*}
where the supremum is over all $D$ in ${\cal D}_X$ and $g$ in $G$.
When $g(x_m) \neq \star$ and 
\[
B(((x_1,g(x_1)),...,(x_{m-1},g(x_{m-1}))),x_m) \neq g(x_m),
\]
we say that $B$ {\em makes a mistake}.

Extend the definition of VC-dimension
to say that the VC-dimension
$\VCdim(G)$ of a set $G$ of functions from $X$ to $\{ 0,\star,1 \}$
% is the largest $d$ for which there exists $(x_1,\ldots,x_d)$ in
% $X^d$ satisfying
% \[
% \{ 0,1 \}^d \subseteq \{ (g(x_1),...,g(x_d)) : g \in G \}.
% \]
is
\[
\max \{ d : \exists x_1,...,x_d \in X,\;
                              \{ 0,1 \}^d \subseteq
                              \{ (g(x_1),...,g(x_d)) : g \in G \} \}.
\]

First, we will make use of the following well-known lemma, whose
application is usually referred to as the ``permutation trick''
(see \cite{HLW}).  It formalizes the idea that, when $m$
points are chosen independently at random, then any permutation of
a certain sequence of points is equally likely to have been chosen.
\begin{lemma}
\label{l:perm}
Choose $m \in \N$, a distribution $D$ on $X$,
and a random variable $\phi$ defined on $X^m$.  
Let $U$ be the uniform distribution on the permutations of
$\{ 1,...,m \}$. Then
\[
\int \phi(x) D^m(x) 
\leq \sup_{(x_1,...,x_m) \in X^m} \int \phi(x_{\sigma(1)},...,x_{\sigma(m)})
                                          U(\sigma).
\]
\end{lemma}

We will make use of the following result of Haussler, Littlestone
and Warmuth.
\begin{lemma}[\cite{HLW}]
\label{l:hlw}
For any $F \subseteq \{ 0,1 \}^{X}$ (note the absence of ``$\star$''),
there is a CWDC strategy $A^\mathrm{one-inc}$ such that for any
points $x_1,...,x_m$, if $U$ is the uniform distribution
over permutations of $\{ 1,...,m\}$, for any $f \in F$,
the probability under $U$ of a permutation $\sigma$ for which
\[
A^\mathrm{one-inc}(((x_{\sigma(1)},f(x_{\sigma(1)})),
        \ldots,(x_{\sigma(m-1)},f(x_{\sigma(m-1)}))),x_{\sigma(m)})
        \not = f(x_{\sigma(m)})
\]
% differs from $f(x_{\sigma(m)})$ is no more than $\VCdim(F)/m$.
is no more than $\VCdim(F)/m$.
\end{lemma}

In the next lemma, we apply a generalization of this result to
give a general upper bound for the CWDC model.
\begin{lemma}
\label{l:dontcare}
Choose $G \subseteq \{ 0,\star,1 \}^X$ and $m \in \N$.  Then
\[
M(G,m) \leq \VCdim(G)/m.
\]
\end{lemma}
{\bf Proof}:  Define a strategy $B$ as follows.
Suppose $B$ is given 
\[
z = (((x_1,y_1),...,(x_{m-1},y_{m-1})),x_m)
\]
as input.  Let $I_z = \{ i \leq m-1 : y_i \neq \star \}$.  Let
$G_z = \{ g \in G : \forall i \in (I_z \cup \{ m \}),  g(x_i) \neq \star \}$.
Note that the restrictions of the functions
of $G_z$ to $\{ x_i : i \in I_z \cup \{ m \} \}$
form a set of $\{ 0,1 \}$-valued functions of VC-dimension at most 
$\VCdim(G)$.  If $G_z$ is empty, $B$ predicts arbitrarily and
doesn't make a mistake, since this implies that it is
certain that $g(x_m) = \star$, where $g$ is the function being learned.
If $G_z$ is non-empty, and $i_1,...,i_k$ are the elements of
$I_z$ in increasing order, then $B$ applies the strategy
from Lemma~\ref{l:hlw} (the one-inclusion graph algorithm)
for learning $G_z$, using $(x_{i_1},y_{i_1}),...,(x_{i_k},y_{i_k}),x_m$
as an input.

Applying Lemma~\ref{l:perm}, we have that for any $g \in G$, the
probability, with respect to $m$ independent random draws from some fixed
distribution, that $B$ makes a mistake on the $m$th prediction, is at most
the maximum, over $x_1,...,x_m$, of the same probability with respect to a
uniformly randomly chosen permutation of $x_1,...,x_m$.

Fix an arbitrary target function
$g$ and sequence $x_1,...,x_m$ of elements of $X$.  
We wish to bound the probability, over a uniformly randomly
chosen permutation $\sigma$ of $x_1,...,x_m$, that $B$ makes a mistake
on the last element, given examples for the first $m-1$.
Let $J = \{ i \leq m : g(x_i) \neq \star \}$
(note that the $m$th element is included in the definition of $J$,
but wasn't in $I_z$ above).
Let $k = |J|$.

Let $E$ be the event that the permutation $\sigma$ moves
one of $\{
i : g(x_i) \neq \star \}$ to be the last, i.e., has $\sigma^{-1}(m) \in J$.
Conditioned on $E$, any order of the elements of $J$ is equally likely, and
furthermore, clearly for any pair of inputs $z_1, z_2$ generated from one
of these permutations in the obvious way, $G_{z_1}$ and $G_{z_2}$ are the
same.  Therefore, by the definition of $B$ and 
Lemma~\ref{l:hlw}, the
probability that $B$ makes a mistake given that the permutation moves one
of $\{ i : g(x_i) \neq \star \}$ to be the last is at most 
$\VCdim(G)/k$.
Next, clearly the probability that $B$ makes a mistake given that the
permutation doesn't send an element of $J$ to be last is $0$, since this
means a $\star$ is last.
Therefore if ``$\Pr$'' is with respect to a random permutation, and
``$\mbox{mistake}$'' is the event that $B$ makes a mistake on
the $m$th element of $X$ when given examples of the first $m-1$, then
\begin{eqnarray*}
\Pr(\mbox{mistake}) 
%    & = &
%       \Pr(\mbox{mistake} | E) \Pr(E)
%       +\Pr(\mbox{mistake} | \neg E) \Pr(\neg E) \\
   & = &
        \Pr(\mbox{mistake} | E) \Pr(E) \\
   & \leq &
        (\VCdim(G)/k) (k/m) \\
   & = &
        \VCdim(G)/m.
\end{eqnarray*}
This completes the proof.  \qed

% We will also make use of the following trivial lemma.
% \begin{lemma}
% \label{l:vcdim}
% Choose $F \subseteq [0,1]^X$, $r\in[0,1]$, and $\gamma >0$.
% Then
% \[
% \VCdim(\psi_{r,\gamma}(F)) \leq \fatV_F (\gamma).
% \]
% \end{lemma}

\noindent{\bf [Warning: The following proof is invalid. See the preceding correction.]}

\noindent{\bf Proof} (of Theorem~\ref{t:binfun}):
Let $d = \fatV_F (\gamma)$.
Consider the strategy $A$ defined as follows.  For each
$r \in [0,1]$, define $B_r$ to be the strategy for
learning $\psi_{r,\gamma}(F)$ described in Lemma~\ref{l:dontcare}.
Given input 
\[
z = (((x_1,y_1),...,(x_{m-1},y_{m-1})),x_m), 
\]
define $z_r$ to be 
\[
(((x_1,\psi_{r,\gamma}(y_1)),
        ...,(x_{m-1},\psi_{r,\gamma}(y_{m-1}))),x_m).
\]
Strategy $A$ performs binary search as described in the
following recurrence.  First, $l_1 = 0$ and $u_1 = 1$.
For each $i \in \N$, 
\begin{itemize}
\item if $B_{(l_i+u_i)/2}(z_{(l_i+u_i)/2}) = 1$, then
$b_i = 1$, $l_{i+1} = (l_i+u_i)/2$, and $u_{i+1} = u_i$, and
\item if $B_{(l_i+u_i)/2}(z_{(l_i+u_i)/2}) = 0$, then
$b_i = 0$, $l_{i+1} = l_i$, and $u_{i+1} = (l_i+u_i)/2$.
\end{itemize}
The output of strategy $A$ is then 
$\sum_{i=1}^{\infty} b_i 2^{-i}$, i.e.\ $0.b_1b_2...$ in binary.

First, by a trivial induction, at any given time during the
binary search, the final prediction of $A$ is contained in
$[l_i,u_i]$.
By an equally trivial induction,
if for $j=1,\ldots,i-1$ either
\[
B_{(l_j+u_j)/2}(z_{(l_j+u_j)/2}) = \psi_{(l_j+u_j)/2,\gamma}(f(x_m))
\]
or $\psi_{(l_j+u_j)/2,\gamma}(f(x_m))=\star$,
then $f(x_m) \in [l_i-\gamma,u_i+\gamma]$.

For each positive integer $i$,
let $E_i$ be the event that
$i$ is the smallest number for which
\[
B_{(l_i+u_i)/2}(z_{(l_i+u_i)/2})\neq
\psi_{(l_i+u_i)/2,\gamma}(f(x_m)) 
\mbox{ and } \psi_{(l_i+u_i)/2,\gamma}(f(x_m)) \neq \star.
\]
Let $E_{\infty}$ be the event that there is no such number.
Then
\[
{\bf E}(|A(z)-f(x_m)|) =
{\bf E}(|A(z)-f(x_m)|\; | \; E_{\infty}) \Pr (E_{\infty}) 
% \hspeq \hspcont
+ \sum_{j\in\N}
                {\bf E}(|A(z)-f(x_m)|\; | \; E_j)
                        \Pr (E_j).
\]
(Here we use the convention that, for each $j$, if
$\Pr(E_j) = 0$, then ${\bf E}(|A(z)-f(x_m)|\; | \; E_j) \Pr (E_j)$
is taken to be $0$.)
Therefore since for any $\hy \in [l,u]$
and $y \in [l-\gamma,u+\gamma]$, it is the case that
$|\hy-y| \leq |l-u|+\gamma$,
we have
\[
{\bf E}(|A(z)-f(x_m)|)
  \leq \gamma \Pr (E_{\infty})
        + \sum_{j\in\N}
                (1/2^{j-1} + \gamma)\Pr (E_j).
\]
Applying Lemma~\ref{l:dontcare}, and the fact that
$\VCdim(\psi_{r,\gamma}(F)) \leq \fatV_F (\gamma)$ for all
$r\in[0,1]$, we get
\begin{eqnarray*}
{\bf E}(|A(z)-f(x_m)|)
  & \leq &
    \gamma      + \sum_{j\in\N}
                (1/2^{j-1}) (d/m)  \\
  & \leq &
    \gamma      + 2d/m.
\end{eqnarray*}
This completes the proof. \qed

The following theorem
shows that Theorem~\ref{t:binfun} cannot be improved in
general by more than a constant factor, and that the constant
on the $\gamma$ term is best possible.  The proof
uses techniques due to
Ehrenfeucht, Haussler, Kearns and Valiant \cite{ehkv},
Haussler, Littlestone and Warmuth \cite{HLW}, and Simon \cite{simon}.
\begin{theorem}
\label{t:binfun.low}
There exists $c$ such that
for all sufficiently small $\gamma \geq 0$, and all
sufficiently large $d, m \in \N$, there is an
$X$, and $F \subseteq [0,1]^X$ such that $\fatV_F(\gamma) = d$ and
\[
{\cal L}(F,m) \geq \max \left\{ c \frac{\fatV_F (\gamma)}{m}, \gamma \right\}.
\]
\end{theorem}
{\bf Proof}:  Consider the class $F$ of all
functions $f$ from $\N$ to $[0,1]$ such that
$|f^{-1}([2 \gamma,1])| \leq d$.  Clearly, $\fatV_F(\gamma) = d$.

We begin by proving the first term.  Consider the distribution $D$ on 
$\{ 1,...,d \}$ where $D(1) = 1-(d-1)/m$ and $D(2) = ... = D(d) = 1/m$.
Clearly, $F$
contains all functions from $\{ 1,...,d\}$ to $\{ 0,1 \}$.  For the
remainder of the proof of the first term, let us assume that $\{ 1,...,d\}$
is the entire domain, and that $F$ consists exactly of those functions from
$\{ 1,...,d \}$ to $\{ 0,1 \}$.  For each $b \in \{ 0,1 \}^d$ define $f_b
\in F$ by $f_b(i) = b_i$.

Fix $u_1,...,u_m \in \{ 1,...,d \}$.
Notice that 
\[
\hy_m = A((u_1,f_b(u_1)),...(u_{m-1},f_b(u_{m-1})),u_m)
\]
is a function only of those components $b_i$ of $b$ for
which $i \in \{ u_1,...,u_{m-1} \}$.
Suppose we choose $b$
according to the uniform distribution over $\{ 0, 1\}^d$.
Choose $i \not \in \{ u_1,...,u_{m-1} \}$,
and $(c_1,...,c_{m-1}) \in \{ 0, 1\}^{m-1}$.
Then, by the independence of the choice of
$b_i$ from that of the other components,
the expectation of $|\hy_m - f_b (i)|$, 
given that $b_{u_1} = c_1,...,b_{u_m} = c_m$, is
\[
(1- \hy_m)/2
 + \hy_m/2,
\]
which, for any value of $\hy_m$, is $1/2$.
Since this is true independent of $c_1,...,c_{m-1}$, for any
$i \not\in \{ u_1,...,u_{m-1} \}$, the expected value of the
error of $A$ on $i$ is at least $1/2$.  

Now, suppose $u_1,...,u_m$ are chosen independently at random
according to $D$ as well.  Then the expectation of $| \hy_m - f_b(u_m)|$ is
at least $1/2$ times the probability that 
$u_m \not\in \{ u_1,...,u_{m-1} \}$.  This probability has
been shown to be $\Omega(d/m)$ \cite{HLW}, and therefore, the
expectation of $| \hy_m - f_b(u_m)|$ over the random choice of
the $u_i$'s and $b$ is $\Omega(d/m)$, which implies there exists
$b$ such that for that fixed $b$, the expectation 
of $| \hy_m - f_b(u_m)|$ only over the random choice of
the $u_i$'s is $\Omega(d/m)$, which completes the proof of
the first term.

The proof of the second term is similar.  Choose $m \in \N$.
Choose a small $\kappa > 0$, and a large $d \in \N$.
Suppose the elements of the domain are chosen according to the
uniform distribution on $\{ 1,...,d \}$, and suppose the
function to be learned is chosen uniformly from the set
of functions from $\{ 1,...,d \}$ to $\{ 0, 2 (\gamma - \kappa) \}$.  
By arguing as above, we can see that the expectation of
\[
|f(x_m) - A((x_1,f(x_1)),...(x_{m-1},f(x_{m-1})),x_m)|,
\]
given that $x_m \not\in \{ x_1,...,x_{m-1} \}$, is at
least $(1/2) (2 (\gamma - \kappa)) = \gamma - \kappa$.
Furthermore, the probability that $x_m \not\in \{ x_1,...,x_{m-1} \}$
is at least $1-(m-1)/d$.  Therefore, the expected error is at
least $(\gamma - \kappa)(1-(m-1)/d)$, and since $\kappa$ can be
made arbitrarily small, and $d$ can be made arbitrarily large, this
completes the proof.
\qed

The following corollary shows that finiteness of $\fatV_F$ at a scale
just below the desired prediction error is sufficient,
and that no larger scale will suffice in general.

\begin{corollary}
Suppose $\epsilon>0$.

For a set $F$ of functions from $X$ to $[0,1]$,
if there is an $\alpha>0$ with $\fatV_F(\epsilon-\alpha)<\infty$,
then for sufficiently large $m$,
${\cal L}(F,m) <\epsilon$.

Moreover, there is a set $F$ such that $\fatV_F(\epsilon)=\infty$
and, for all $\alpha>0$, $\fatV_F(\epsilon+\alpha)=0$,
but ${\cal L}(F,m) \ge\epsilon$ for all $m$.
\end{corollary}

The proof of the sufficient condition follows on substituting
$\gamma=\epsilon-\alpha$ in Theorem~\ref{t:binfun}.
The converse result is exhibited by
the class $F$ of all functions from $\N$ to $\{0,2\epsilon\}$,
using similar techniques to the proof of
Theorem~\ref{t:binfun.low}.

In later sections,
we investigate the scale at which the dimensions $\fat$ and $\fatV$
need to be finite for agnostic learnability.
The following result shows that precise bounds on this scale are
important,
since a constant factor gap in the scale can lead to an arbitrarily
large gap in the sample complexity bounds.

\begin{proposition}
For any non-increasing function $\phi$ from $(0,1/2]$ to
$\N\cup\{0,\infty\}$,
there is a function class $F_\phi:\N\to[0,1]$ that satisfies
$\fatV_F(\gamma)=\phi(\gamma)$ for all $\gamma$ in $(0,1/2]$.
\end{proposition}

\begin{proof}
Let $\{A_{d,n}:d\in\N\cup\{\infty\}, n\in\N\}$ be a partition of
$\N$, with $|A_{d,n}|=d$ for $d,n\in\N$, and $A_{\infty,n}$
countably infinite for all $n\in\N$.

Fix $d\in\N\cup\{\infty\}$, and
consider the set $S_d=\phi^{-1}(d)\subset[0,1/2]$.
If $S_d$ is empty, let $F_{d,n}=\emptyset$ for $n\in\N$.
Otherwise, since $\phi$ is non-increasing, $S_d$ is an interval.
Suppose $r=\sup S_d$.
There are two cases:

\noindent
{\bf Case 1:} $r$ is in $S_d$.\\
Let $F_{d,1}$ be the set of all functions $f$ satisfying
$f(x) \in \{1/2-r,1/2+r\}$ if $x\in A_{d,1}$, and $f(x)=0$ otherwise.
Let $F_{d,n}=\emptyset$ for $n>1$.

\noindent
{\bf Case 2:} $r$ is not in $S_d$.\\
For $n\in\N$, let $F_{d,n}$ be the set of all functions $f$
satisfying
$f(x) \in \{1/2-r(1-1/n),1/2+r(1-1/n)\}$ if $x\in A_{d,n}$
and $f(x)= 0$ otherwise.

Let
\[
F=\bigcup\left\{F_{d,n}:n\in\N,d\in\N\cup\{\infty\}\right\}.
\]

For any $d$ in $\N\cup\{\infty\}$,
the sets $F_{d,n}$ ensure that for all $\gamma\in\phi^{-1}(d)$,
$\fatV_F(\gamma)\ge d$.
Clearly, any set of points in $\N$ that has nonempty intersection
with two distinct $A_{d,n}$'s cannot be $\gamma$-shattered
for any $\gamma > 0$,
which implies that the reverse inequality is also true.
The case $d=0$ is trivial, and hence
$\fatV_F(\gamma)=\phi(\gamma)$ for all
$\gamma\in\phi^{-1}\left(\N\cup\{0,\infty\}\right)$.
\end{proof}

% 4pack.tex

\section{Packing number bounds}
\label{s:pack}

In this section, we prove two new bounds on ${\cal M}(\epsilon,S)$.
One uses $\fatV$, and is proved using Theorem~\ref{t:binfun},
together with techniques from~\cite{haussler-sphere,benedek}.
The second bound uses $\fat$, and is proved through a refinement
% took out similar - implied since it is a refinement
of a proof in~\cite{abch}.

For a set $X$, and $F \subseteq [0,1]^X$, define 
\[
m_{\cal L}(\epsilon,F) = \min \{ m \in\N: {\cal L}(F,m) \leq \epsilon \}.
\]
The following bound on $m_{\cal L}(\epsilon,F)$ follows immediately
from Theorem~\ref{t:binfun}.
\begin{lemma}
\label{l:fatV.predsamp}
Choose $X$, $F \subseteq [0,1]^X$, and $\alpha, \epsilon > 0$.
Assume $\fatV_F(\epsilon - \alpha)\ge 1$.  Then
\[
m_{\cal L}(\epsilon,F) \leq 2 \fatV_F(\epsilon - \alpha)/\alpha.
\]
\end{lemma}

For $m \in \N, x = (x_1,...,x_m) \in X^m$, and $\fun{f}{X}{[0,1]}$,
define
\[
\sam(x,f) = ((x_1,f(x_1)),...,(x_m,f(x_m))).
\]
We will
also make use of the following, which is implicit in the
work of Haussler, et al~\cite{HLW}.
\begin{lemma}
\label{l:pred.learn}
Choose $X$, $F \subseteq [0,1]^X$.  There is a learner $A$ such that
for all $f \in F$, for any distribution $D$ on $X$, for all $m \in \N$,
\[
\int \left(\int |(A(\sam(x,f)))(u) - f(u)|\; dD(u)\right)\;dD^{m-1}(x)
\]
is no more than ${\cal L}(F,m)$.
\end{lemma}

We apply these in the following.  In addition to Theorem~\ref{t:binfun},
the proof uses ideas due to Haussler~\cite{haussler-sphere} and Benedek and
Itai~\cite{benedek}.
\begin{lemma}
\label{l:fatv.packbound}
Choose $0<\epsilon<1$, $b \in \N$ and $0<\alpha <\epsilon/4$.
Let $B = Q_{1/b}([0,1])$.  Choose
$m \in \N$, and let $S \subseteq B^m$.  
Set $d = \fatV_S (\epsilon/2 - \alpha)$.
Then if $d\ge 1$,
\[
{\cal M}(\epsilon,S) \leq \frac{\epsilon}{2\alpha}(b+1)^{4d/\alpha}.
\]
\end{lemma}
{\bf Proof}:  For each $v \in S$, define
$\fun{f_v}{\{ 1,...,m\}}{[0,1]}$ by $f_v(i) = v_i$, and
define $F = \{ f_v : v \in S \}$.  Let $D$ be the uniform
distribution on $\{ 1,...,m \}$.  Then for all $v,w \in S$,
\begin{equation}
\label{e:ellone}
\ell_1 (v,w) = \int |f_v(u) - f_w (u)|\; dD(u).
\end{equation}
Define $\ell_1 (f,g) = \int |f(u) - g (u)|\; dD(u)$.
Let $m_0 = m_{\cal L}(\epsilon/2-\alpha,F)$.  
Then, by Lemma~\ref{l:pred.learn},
there is a learner $A$ such that for all $f \in F$,
\begin{equation}
\label{e:ind.small}
\int \ell_1(A(\sam(x,f)),f)\;dD^{m_0}(x)
      \leq \epsilon/2 - \alpha.
\end{equation}
Choose an $\epsilon$-separated subset $T$ of
${\cal M}(\epsilon,S)$ elements of $S$.
Then by (\ref{e:ind.small}), we have
\[
\sum_{v \in T}
\int \ell_1(A(\sam(x,f_v)),f_v)\;dD^{m_0}(x)
      \leq (\epsilon/2 - \alpha) |T|,
\]
and hence
\begin{equation}
\label{e:sum.up}
\int \sum_{v \in T}
\ell_1(A(\sam(x,f_v)),f_v)\;dD^{m_0}(x)
      \leq (\epsilon/2 - \alpha) |T|.
\end{equation}
Fix $x \in \{1,\ldots,m\}^{m_0}$.  For any set $T' \subseteq T$ such that for
all $f_1$ and $f_2$ in $T'$ it is the case that $\sam(x,f_1) = \sam(x,f_2)$,
since $T'$ is $\epsilon$-separated, the triangle inequality implies that
\[
\ell_1(A(\sam(x,f)),f) < \epsilon/2
\]
for no more than one $f$ in $T'$.
% For any $v,w \in T$ such that
% $\sam(x,f_v) = \sam(x,f_w)$, since $\ell_1(f_v,f_w) \geq \epsilon$,
% by the triangle inequality, either 
% \[
% \ell_1(A(\sam(x,f_v)),f_v) \geq \epsilon/2,
% \]
% or
% \[
% \ell_1(A(\sam(x,f_w)),f_w) \geq \epsilon/2.
% \]
Therefore,
if we let $f_v(x) = (f_v(x_1),\ldots,f_v(x_{m_0}))$,
we have
\[
\begin{array}{l}
\sum_{v \in T}
\ell_1(A(\sam(x,f_v)),f_v) \\
\hspace{1cm}
 = \sum_{l \in B^{m_0}} \sum_{v \in T, f_v(x) = l}
             \ell_1(A(\sam(x,f_v)),f_v) \\
\hspace{1cm}
 \geq \sum_{l \in B^{m_0}} 
          (\epsilon/2)(|\{ v \in T : l = f_v(x)\}|-1) \\
\hspace{1cm}
 \geq (\epsilon/2)(|T| - (b+1)^{m_0}).
\end{array}
\]
This inequality, together with (\ref{e:sum.up}), implies
\[
(\epsilon/2)(|T| - (b+1)^{m_0}) \leq (\epsilon/2 - \alpha) |T|.
\]
Solving for $|T|$, and recalling that $|T| = {\cal M}(\epsilon,S)$
and $m_0 = m_{\cal L}(\epsilon/2-\alpha,F)$,
gives
\[
{\cal M}(\epsilon,S) \leq \frac{\epsilon}{2\alpha}(b+1)^{m_{\cal
L}(\epsilon/2-\alpha,F)}.
\]
Applying Lemma~\ref{l:fatV.predsamp}
completes the proof. \qed

Next, we give a new bound on ${\cal M}(\epsilon,S)$ in
terms of $\fat_S$.  
Its proof is based on that of a
corresponding lemma in~\cite{abch} which dealt with the $\ell_{\infty}$
norm.
\begin{lemma}
\label{l:fat.packbound}
Choose $\epsilon > 0$.  Choose $b \in \N$, $b > 4/\epsilon$.  
Let $B = Q_{1/b}([0,1])$.  Choose
$m \in \N$, and let $S \subseteq B^m$.  
Set $d = \fat_S (\epsilon/2 - 2/b)$,
and 
\[
y = \sum_{i=0}^d \comb{m}{i} (1+b)^i.
\]
Then if $m\ge d$,
%PL I've done some calcs for this r.e. the ref's comment, but I
%   think you could redo them quicker than I could typeset them or
%   fax them.
\[
{\cal M}(\epsilon,S) \leq 2
        b^{3 \left(\lfloor \log_2 y \rfloor+1\right)}
        \leq 2 b^{6d\log_2(2bem/d)}.
\]
\end{lemma}
{\bf Proof}:  Fix $m \in \N$.  For each $h \in \N$ let
$t(h)$ be the minimum, over all 
$S \subseteq B^m$ with $|S| \geq h$ that are pairwise $\epsilon$-separated in the
$\ell_1$ norm, of the number of
finite sets $\{ (i_1,r_1),..., (i_k,r_k) \}$ of elements of 
$\{ 1,...,m \} \times B$ which are
$(\epsilon/2 - 2/b)$-fatly shattered by $S$.
(Here we say a set is $\gamma$-fatly shattered if any corresponding
sequence is $\gamma$-fatly shattered.)
Obviously, $t(2) \geq 1$.

Choose an even $h$, and let $S$ be a pairwise
$\epsilon$-separated subset of $B^m$.  Split $S$ arbitrarily into 
$h/2$ pairs.  For each pair 
$v$ and $w$, if 
\[
l = \left|\{ i : |v_i-w_i| \geq \epsilon - 1/b \}\right|,
\]
then
$\ell_1(v,w) = \frac{1}{m} \sum_{i=1}^m | v_i - w_i| < \frac{1}{m}
(l + (\epsilon - 1/b) m) = l/m + \epsilon - 1/b$.
% \begin{eqnarray*}
% \ell_1(v,w) 
%       & = & \frac{1}{m} \sum_{i=1}^m | v_i - w_i| \\
%       & < & \frac{1}{m} (l + (\epsilon - 1/b) m) \\
%       & = & l/m + \epsilon - 1/b.
% \end{eqnarray*}
But $\ell_1(v,w) \geq \epsilon$, thus $l/m + \epsilon - 1/b > \epsilon$,
which implies $l \geq m/b$.  Thus each pair
$(v,w)$ has at least $m/b$ indices $i$ such that
$| v_i - w_i | \geq \epsilon - 1/b$.  Applying the pigeonhole
principle, there is some index $i_0$ such that
$h/(2b)$ pairs $v$ and $w$ have $|v_{i_0} - w_{i_0}| 
\geq \epsilon - 1/b$.
Again, by the pigeonhole principle, there are at least
$h/(b^2(b-1))\ge h/b^3$ such pairs for which the pair
$\{ v_{i_0}, w_{i_0} \}$ is the same.

This implies that there are two subsets $S_1$ and $S_2$ of $S$
having at least $h/b^3$ elements each, 
and $y_1,y_2 \in B$ with $|y_1 - y_2 | \geq \epsilon - 1/b$, such that,
for each $v \in S_1$, $v_{i_0} = y_1$, and each
$v \in S_2$, $v_{i_0} = y_2$.  Obviously, any two points
in $S_1$, respectively $S_2$ are $\epsilon$-separated, thus
$S_1$ $(\epsilon/2 - 2/b)$-fatly shatters at least
$t (\ceil{h/b^3})$ sets, as does $S_2$.
If the same set $\left\{(i_1,r_1),...,(i_k,r_k)\right\}$
is shattered by both, then so is 
\[
\left\{\left(i_0,Q_{1/b}\left(\frac{v_{i_0}+w_{i_0}}{2}\right)\right),
(i_1,r_1),...,(i_k,r_k)\right\}.
\]
Thus, 
% \[
$t(h) \geq 2 t\left(\ceil{h/b^3}\right)$.
% \]
Since $t(2)\ge 1$,
by induction, for all $k$, $t(2b^{3k})\ge 2^k$,
and therefore
\[
% $
t\left(2b^{3\left(\floor{\log_2 y}+1\right)}\right) > y.
% $
\]
However, as argued in~\cite{abch}, there are only $y$
% suitable sequences
sets of at most $d$
% \[
% (i_1,r_1),...,(i_k,r_k)
% \]
elements of $\{1,\ldots,m\}\times B$.
% of length no more than $d$.
% for which $k \leq d$.
But, by the definition of $t$,
the fact that
\[
% $
t(2b^{3\left(\floor{\log_2 y}+1\right)}) > y
% $
\]
implies that
any $\epsilon$-separated subset of
$2 b^{3 \left(\floor{\log_2 y}+1\right)}$
elements must $(\epsilon/2 - 2/b)$-fatly shatter more than $y$ 
sets, and therefore a set of length at least $d+1$.
Thus, no such subset can have
$\fat_S(\epsilon/2-2/b)$ at most $d$.
Taking the contrapositive completes the proof of the first
inequality in the lemma.

The second inequality is obtained by bounding $y$ using
Sauer's lemma (see, for example,~\cite{behw}).
\qed

% 5ag.tex

\section{Sample complexity bounds}
\label{s:sample}

In this section, we apply the bounds of the previous section
to upper bound the sample size necessary for agnostic
learnability, and for uniformly good estimates of the
expectations of a set of random variables.  We start with
the latter.

\begin{theorem}
\label{t:sample}
Choose $X$, and a set $F$ of functions from $X$ to $[0,1]$.

If there is a $\kappa > 0$ such that for all $\epsilon > 0$,
$\fat_F((1/4-\kappa)\epsilon)$
is finite, then
\begin{equation}                \label{equation:fatGCbound}
m_{GC,F} (\epsilon,\delta) 
= O\left( 
                        \frac{1}{\epsilon^2}\left(\fat_F((1/4-\kappa)\epsilon)
                                \log^2 \frac{1}{\epsilon} + \log \frac{1}{\delta} 
                                                \right) \right).
\end{equation}
If there is a $\kappa > 0$ such that for all $\epsilon > 0$,
$\fatV_F((1/4-\kappa)\epsilon)$
is finite, then
\begin{equation}                \label{equation:fatVGCbound}
m_{GC,F} (\epsilon,\delta) 
 = O\left(\frac{1}{\epsilon^2}\left(
                \frac{1}{\epsilon}\fatV_F((1/4-\kappa)\epsilon)
                        \log\frac{1}{\epsilon}
                +\log\frac{1}{\delta}\right)\right).
\end{equation}
\end{theorem}

Before sketching the proof of Theorem~\ref{t:sample}, we establish some
lemmas.
The first is Hoeffding's inequality (see~\cite{pollard},
Appendix B).
\begin{lemma}
\label{l:hoeff}
Choose $a < b$, $X$. Let $D$ be a probability
distribution on $X$, and let $f_1,...,f_m$ be independent
random variables taking values in $[a,b]$.
Then the probability under $D^m$ of a sequence
$(x_1,...,x_m)$ for which
\[
\left|\left( \frac{1}{m} \sum_{i=1}^m f_i(x_i) \right)
- \left(\frac{1}{m} \sum_{i=1}^m \int f_i(x) D(x)\right) \right| >
\epsilon
\]
is no more than $2 e^{-2 \epsilon^2 m /(b-a)^2}$.
% \[
% D \left\{ (x_1,...,x_m) : \left|\left( \frac{1}{m} \sum_{i=1}^m
% f_i(x_i) \right)
%           - \left(\frac{1}{m} \sum_{i=1}^m \int f_i(x) D(x)\right)
% \right| > \epsilon \right\}
%      \leq 2 e^{-2 \epsilon^2 m /(b-a)^2}.
% \]
\end{lemma}

The following is a restatement of Theorem 8 of Chapter II
of~\cite{pollard}.
\begin{lemma}[\cite{pollard}]
\label{l:symlemlem}
Suppose $X$ and $U$ are sets, $D$ is a probability distribution on
$X$, and $\fun{\Phi}{X \times U}{[0,1]}$ and
$\fun{\Psi}{X \times U}{[0,1]}$ are functions for which
$\Phi(\cdot,u_1)$ and $\Psi(\cdot,u_2)$ are independent random
variables for all $u_1$ and $u_2$ in $U$.
Suppose there exist constants $\beta, \alpha > 0$
such that for all $u \in U$, 
$D \{ x : |\Psi(x,u)| \leq \alpha \} \geq \beta$.  Then for all
$\epsilon > 0$,
\[
D \{ x : \sup_u \Phi(x,u) > \epsilon \}
   \leq \frac{1}{\beta} D \{ x : \sup_u |\Phi(x,u) - \Psi(x,u)| >
\epsilon-\alpha \}.
\]
\end{lemma}

% This is applied in the following.
These are applied in the following.

\begin{lemma}
\label{l:sym}
Choose a set $X$, and a set $F \subseteq [0,1]^X$.  Choose $\epsilon > 0$,
$0 < \alpha < \epsilon$ and $m \in \N$.  Then
\[
\begin{array}{l}
D^m \{ (x_1,...,x_m) : \exists f \in F  
                        \left|\left( \frac{1}{m} \sum_{i=1}^m f(x_i) \right)
                          - \int f(x) \; dD(x)\right| > \epsilon \} \\
\hspeq  \leq
                \frac{1}{1-2 e^{-2 \alpha^2 m}}
D^{2m} \{ (x_1,...,x_m,y_1,...,y_m) : \exists f \in F 
                        \left|\left( \frac{1}{m} \sum_{i=1}^m f(x_i) \right)
                         - \left( \frac{1}{m} \sum_{i=1}^m f(y_i) \right)\right|
                         > \epsilon - \alpha \}.
\end{array}
\]
\end{lemma}
{\bf Proof}:  In Lemma~\ref{l:symlemlem}, set
$X$ to be this lemma's $X^{2m}$, $U$ to be $F$,
$\Phi$ to be defined by 
\[
\Phi((x_1,...,x_m,y_1,...,y_m),f) 
 = \frac{1}{m} \sum_{i=1}^m f(x_i) - \int
f(x) \; D(x)
\]
and $\Psi$ by
\[
\Psi((x_1,...,x_m,y_1,...,y_m),f) 
 = \frac{1}{m} \sum_{i=1}^m f(y_i) - \int
f(x) \; D(x).
\]
% Applying the standard Hoeffding bound (see \cite[Appendix B]{pollard}), 
Applying the standard Hoeffding bound (Lemma~\ref{l:hoeff}),
we get for all $f \in F$
\[
D^{2m} \{ (x_1,...,x_m,y_1,...,y_m) : |\Psi((x_1,...,x_m,y_1,...,y_m),f)| 
                                        \leq \alpha \} 
             \geq 1 - 2 e^{-2 \alpha^2 m}.
\]
Applying Lemma~\ref{l:symlemlem} completes the proof. \qed

\begin{lemma}
\label{l:twosamp}
Choose $X$, $F \subseteq [0,1]^X$.  Let $D$ be a probability distribution over
$X$. 
Choose $0 < \alpha, \epsilon < 1$ with $\alpha<\epsilon/2$.
Then
\[
\begin{array}{l}
D^{2m} \{ (x_1,...,x_m,y_1,...,y_m) : \exists f \in F 
             \left|\left( \frac{1}{m} \sum_{i=1}^m f(x_i) \right)
                  - \left( \frac{1}{m} \sum_{i=1}^m f(y_i) \right)\right|
                     > \epsilon \} \\
\hspeq \leq 2 \left(\sup_{\xi \in X^{2m}} 
                       {\cal N}(\epsilon/2 - \alpha,\rest{F}{\xi})\right)
                            e^{- 2 \alpha^2 m}.
\end{array}
\]
\end{lemma}
{\bf Proof}:  Let $U$ be the uniform distribution over $\{ -1,1 \}$.
Then by symmetry,
\begin{equation}
\label{e:perm}
\begin{array}{l}
D^{2m} \{ (x_1,...,x_m,y_1,...,y_m) : \exists f \in F 
                        \left|\left( \frac{1}{m} \sum_{i=1}^m f(x_i) \right)
                         - \left( \frac{1}{m} \sum_{i=1}^m f(y_i) \right)\right|
                         > \epsilon \} \\[3pt]
 = (D^{2m}\times U^m) \{ (x_1,...,x_m,y_1,...,y_m), (u_1,...,u_m) : \exists f \in F 
                        \left|\frac{1}{m} 
                        \sum_{i=1}^m u_i (f(x_i)- f(y_i)) \right|
                         > \epsilon \} \\[3pt]
 \leq \sup_{(x_1,...,x_m,y_1,...,y_m)} U^m \{ (u_1,...,u_m) : \exists f \in F 
                        \left|\left( \frac{1}{m} 
                        \sum_{i=1}^m u_i (f(x_i)- f(y_i)) \right)\right|
                         > \epsilon \}.
\end{array}
\end{equation}

Fix $\sigma = (x_1,...,x_m,y_1,...,y_m)$.  

Choose $f \in F$.  Suppose $\fun{g}{X}{[0,1]}$ had
\[
\frac{1}{2m} \sum_{i=1}^m (|g(x_i)-f(x_i)| + |g(y_i)-f(y_i)|) 
        \leq \epsilon/2 - \alpha,
\]
and that
\[
                \left|\frac{1}{m} 
                        \sum_{i=1}^m u_i (f(x_i)- f(y_i)) \right|
                         > \epsilon.
\]
Then
\begin{eqnarray*}
\left|\frac{1}{m} \sum_{i=1}^m u_i (g(x_i)- g(y_i)) \right|
        & = &   \left|\frac{1}{m} \sum_{i=1}^m u_i (g(x_i) - f(x_i) +
                        f(x_i) - g(y_i) + f(y_i) - f(y_i)) \right| \\
        & = &   \left|\frac{1}{m} \sum_{i=1}^m u_i (f(x_i) - f(y_i))
                        +\frac{1}{m} \sum_{i=1}^m u_i (g(x_i) - f(x_i)- g(y_i)
                        + f(y_i)) \right| \\
        & \geq& \left|\frac{1}{m} \sum_{i=1}^m u_i (f(x_i) - f(y_i))\right|
                        -\left| \frac{1}{m} \sum_{i=1}^m (|g(x_i) - f(x_i)| +
                        |g(y_i) - f(y_i)|) \right| \\
        & > &   \epsilon - (\epsilon - 2 \alpha) \\
        & = &   2 \alpha.
\end{eqnarray*}
So if $T$ is a (minimum-sized) 
$(\epsilon/2 -\alpha)$-cover of $\rest{F}{\sigma}$, then
\begin{equation}
\label{e:covsuff}
\begin{array}{l}
U^m \{ (u_1,...,u_m) : \exists f \in F 
                        \left|\left( \frac{1}{m} 
                        \sum_{i=1}^m u_i (f(x_i)- f(y_i)) \right)\right|
                         > \epsilon \} \\[3pt]
  \leq
 \sum_{v \in T} U^m \{ (u_1,...,u_m) : 
                        \left|\left( \frac{1}{m} 
                        \sum_{i=1}^m u_i (v_i- v_{m+i}) \right)\right|
                         > 2 \alpha \}.
\end{array}
\end{equation}

Fix $v \in X^{2m}$.  Then $u_1 (v_1 - v_{m+1}),..., u_m (v_m - v_{2m})$
form a sequence of independent $[-1,1]$ random variables with zero mean.
Applying Hoeffding's inequality, we get
\[
U^m \left\{ (u_1,...,u_m) : 
                        \left|\left( \frac{1}{m} 
                        \sum_{i=1}^m u_i (v_i- v_{m+i}) \right)\right|
                         > 2 \alpha  \right\}
        \leq 2 e^{-2 \alpha^2 m}.
\]
Combining this with 
\[
|T| \leq \sup_{\sigma \in X^{2m}}
{\cal N}(\epsilon/2 - \alpha,\rest{F}{\sigma}),
\]
(\ref{e:covsuff}), and (\ref{e:perm}) completes the proof. \qed

% We will make use of the following simple lemma.
\begin{lemma}
\label{l:quantok}
Choose $S \subseteq [0,1]^m$, $\epsilon > 0$, $\alpha < \epsilon/2$.  Then
\[
{\cal N} (\epsilon,S) \leq {\cal N}(\epsilon-\alpha,Q_{\alpha}(S)).
\]
\end{lemma}
{\bf Proof}:  Choose $v,w \in [0,1]^m$.  
\begin{eqnarray*}
\ell_1 (v,Q_{\alpha}(w))
        & = &   \frac{1}{m} \sum_{i=1}^m | v_i - \alpha \floor{w_i/\alpha}| \\
        & = &   \frac{1}{m} \sum_{i=1}^m | (v_i - w_i)
                           + \alpha(w_i/\alpha- \floor{w_i/\alpha}) | \\
        & \geq& \ell_1 (v,w) - \alpha.
\end{eqnarray*}
Thus $\ell_1 (v,w) \leq \ell_1 (v,Q_{\alpha}(w)) + \alpha$.
Therefore, if some $T$ is an $(\epsilon - \alpha)$-cover of $Q_{\alpha}(S)$,
then $T$ is an $\epsilon$-cover of $S$, completing the proof. \qed

Next, we write down a lemma calculating a useful inverse.
The lemma is proved using the by now standard technique
from \cite{abst90}.
% It seems misleading to cite abst90 for the lemma.
% \begin{lemma}[\cite{abst90}]
\begin{lemma}
\label{l:once.and.for.all.again}
For any $y_1,y_2,y_4,\delta>0$ and $y_3 \geq 1$, if
\[
m \geq \frac{2}{y_4} \left( 
       y_2 \ln \left(\frac{2 y_2 y_3}{y_4} \right)
       + \ln \frac{y_1}{\delta} \right),
\]
then
\[
y_1 \exp (y_2 \ln (y_3 m) - y_4 m) \leq \delta.
\]
\end{lemma}
{\bf Proof}:  If $\gamma = y_4/(2 y_2 y_3)$, then
\[
m \geq \frac{2}{y_4} \left( 
       y_2 \ln \left(\frac{2 y_2 y_3}{y_4} \right)
       + \ln \frac{y_1}{\delta} \right)
\]
implies
\[
\left( 1 - \frac{\gamma y_2 y_3}{y_4} \right) m
        \geq 
       \frac{1}{y_4} \left( 
       y_2 \ln \left(\frac{1}{\gamma} \right)
       + \ln \frac{y_1}{\delta} \right).
\]
Solving for $m$, we get
\[
m \geq 
       \frac{1}{y_4} \left( 
       y_2 \left(\gamma y_3 m + \ln \left(\frac{1}{\gamma} \right)\right)
       + \ln \frac{y_1}{\delta} \right).
\]
Applying the fact~\cite{abst90} that for all $x,\gamma > 0$,
$\ln x +\ln\gamma \le \gamma x$ with $x = y_3 m$, we get
\[
m \geq \frac{1}{y_4} \left( 
       y_2 \ln (y_3 m)
       + \ln \frac{y_1}{\delta} \right).
\]
Solving for $\delta$ completes the proof.
\qed

{\bf Proof} (of Theorem~\ref{t:sample}): Choose a probability
distribution $D$ on $X$.  Let $\alpha = 1/\lceil 1/(\kappa \epsilon)\rceil$.  
Since $\fat_F$ is nonincreasing, 
$\fat_F(\epsilon/4 - \alpha) \leq \fat_F((1/4-\kappa)\epsilon)$.
Let $d = \fat_F
(\epsilon/4-\alpha)$.  By Lemma~\ref{l:sym},
\[
\begin{array}{l}
D^m \{ (x_1,...,x_m) : \exists f \in F 
                        \left|\left( \frac{1}{m} \sum_{i=1}^m f(x_i) \right)
                          - \int f(x) \; dD(x)\right| > \epsilon \} \\[3pt]
\hspace{5pt}    \leq
                \frac{1}{1-2 e^{-2\alpha^2 m}}
D^{2m} \{ (x_1,...,x_m,y_1,...,y_m) : \exists f \in F 
                        \left|\left( \frac{1}{m} \sum_{i=1}^m f(x_i) \right)
                         - \left( \frac{1}{m} \sum_{i=1}^m f(y_i) \right)\right|
                         > \epsilon - \alpha \}.
\end{array}
\]
Applying Lemmas~\ref{l:twosamp} and~\ref{l:quantok} yields
\begin{eqnarray*}
&&
D^m \left\{ (x_1,...,x_m) : \exists f \in F 
                        \left|\left( \frac{1}{m} \sum_{i=1}^m f(x_i) \right)
                          - \int f(x) \; dD(x)\right| > \epsilon \right\} \\
&& \hspace{1in}  \leq 2 (\sup_{\sigma \in X^{2m}} 
        {\cal N}(\epsilon/2 -\alpha), \rest{F}{\sigma})
        \frac{e^{-\alpha^2 m/2}}{1-2 e^{-2\alpha^2 m}}\\
&& \hspace{1in}  \leq 2 (\sup_{\sigma \in X^{2m}} 
        {\cal N}(\epsilon/2-8\alpha/7), Q_{\alpha/7}(\rest{F}{\sigma}))
        \frac{e^{-\alpha^2 m/2}}{1-2 e^{-2\alpha^2 m}}.
\end{eqnarray*}
It is immediate from the definition of $\fat_F$ that
for all $0<\beta<\gamma$,
$\fat_{Q_\beta(F)}(\gamma)\le\fat_F(\gamma-\beta)$,
so
$\fat_{Q_{\alpha/7}(F)}(\epsilon/4-6\alpha/7)\le
\fat_F(\epsilon/4-\alpha)$.
Lemma~\ref{l:fat.packbound} implies that
\[
{\cal N} \left(\epsilon/2-8\alpha/7,
        Q_{\alpha/7}(\rest{F}{\sigma})\right)
\leq
2 \left(\frac{7}{\alpha}\right)^
        {6d\log_2\left(\frac{14em}{\alpha d}\right)}.
\]
% provided $m\ge 2d\log_2\left(28e/(\alpha\ln 2)\right)$.
If $m > \frac{1}{2\alpha^2}\ln 4$,
$1/(1-2e^{-2\alpha^2m})<2$.
In that case,
the probability above is less than
\[
%PL removed exponent 3
% 8\exp\left(\frac{6d}{\ln 2}\ln\frac{14em}{\alpha
% d}\ln\frac{7}{\alpha^3} - \frac{\alpha^2m}{2}\right),
8\exp\left(\frac{6d}{\ln 2}\ln\frac{14em}{\alpha
d}\ln\frac{7}{\alpha} - \frac{\alpha^2m}{2}\right),
\]
which, by Lemma~\ref{l:once.and.for.all.again},
is no more than $\delta$ if
\begin{eqnarray*}
m
  & \ge &
\frac{4}{\alpha^2}\left(\frac{6d}{\ln 2}\ln\frac{7}{\alpha}
\ln\left(\frac{336 e}{\alpha\ln 2}\ln\frac{7}{\alpha}\right)
+\ln\frac{8}{\delta}\right)\\
  & = &
O\left(\frac{1}{\alpha^2}\left(d\log^2\frac{1}{\alpha}+
\log\frac{1}{\delta}\right)\right),
\end{eqnarray*}
which completes the proof of~(\ref{equation:fatGCbound}).

A similar argument gives~(\ref{equation:fatVGCbound}).
In this case,
let $d=\fatV_F(\epsilon/4-\alpha)$.
By Lemmas~\ref{l:sym}, \ref{l:twosamp}, \ref{l:quantok},
and~\ref{l:fatv.packbound},
and the fact that
$\fatV_{Q_{\alpha/5}(F)}(\epsilon/4-4\alpha/5)\le
\fatV_F(\epsilon/4-\alpha)$,
we have
\begin{eqnarray*}
&&
D^m \left\{ (x_1,...,x_m) : \exists f \in F 
                        \left|\left( \frac{1}{m} \sum_{i=1}^m f(x_i) \right)
                          - \int f(x) \; dD(x)\right| > \epsilon \right\} \\
&& \hspace{1in}  \leq 2 \left(\sup_{\sigma \in X^{2m}} 
        {\cal N}(\epsilon/2 -2\alpha), \rest{F}{\sigma}\right)
        \frac{e^{-2\alpha^2 m}}{1-2 e^{-8\alpha^2 m}}\\
&& \hspace{1in}  \leq 2 (\sup_{\sigma \in X^{2m}} 
        {\cal N}(\epsilon/2-11\alpha/5), Q_{\alpha/5}(\rest{F}{\sigma}))
        \frac{e^{-2\alpha^2 m}}{1-2 e^{-8\alpha^2 m}}\\
&& \hspace{1in}  \leq \frac{4\epsilon}{\alpha}
        \left(\frac{5}{\alpha}+1\right)^{16d/\alpha}
        \frac{e^{-2\alpha^2 m}}{1-2 e^{-8\alpha^2 m}},
\end{eqnarray*}
and this quantity is less than $\delta$ when
\begin{eqnarray*}
m       & \ge & \frac{8d}{\alpha^3}\ln\frac{6}{\alpha}+
        \frac{1}{2\alpha^2}\ln\frac{8\epsilon}{\delta\alpha} \\
        & = &   O\left(\frac{1}{\alpha^2}\left(
                \frac{1}{\alpha}\fatV_F(\epsilon/4-\alpha)\log\frac{1}{\alpha}
                +\log\frac{1}{\delta}\right)\right).
\end{eqnarray*}
\qed

To use Theorem~\ref{t:sample} to give sample size bounds for agnostic
learnability, we will consider an algorithm that approximately
minimizes empirical loss.
In this case,
we need to show that a class of associated loss functions is an
$\epsilon$-uniform GC class,
and to do this we relate covering numbers of the loss function class
to covering numbers of the function class.
This lemma is implicit in the analysis of
Natarajan \cite{nat}.\footnote{It is also possible to relate the fat-shattering
functions of these classes directly using Sauer's lemma
(see~\cite{koiran}), but the proof is not as simple and the
result slightly weaker.}

\begin{lemma}
\label{l:losspack}
Suppose that $X$ is a set, $F$ is a class of functions that map
from $X$ to $[0,1]$, $x=(x_1,\ldots,x_m)\in X^m$,
and $z=((x_1,y_1),\ldots,(x_m,y_m))\in\left(X\times[0,1]\right)^m$.
Define the loss function class
\[
% $ 
L_F = \left\{(x,y)\mapsto |f(x)-y|: f\in F\right\}.
% $
\]
Then for any $\epsilon>0$,
${\cal N}\left(\epsilon,L_{\rest{F}{z}}\right)\le
{\cal N}\left(\epsilon,\rest{F}{x}\right)$.
% \[
% {\cal N}\left(\epsilon,L_{\rest{F}{z}}\right)\le
% {\cal N}\left(\epsilon,\rest{F}{x}\right).
% \]
\end{lemma}
\iffalse
{\bf Proof}:
Suppose $T\subseteq[0,1]^m$ is an $\epsilon$-cover of $\rest{F}{x}$.
Then for all $f$ in $F$ there is a $t$ in $T$ with
$\frac{1}{m}\sum_{i=1}^m | t_i-f(x_i)|\le\epsilon$.
% \[
% \frac{1}{m}\sum_{i=1}^m | t_i-f(x_i)|\le\epsilon.
% \]
For this $f$ and $t$,
\[
\frac{1}{m}\sum_{i=1}^m \left| ( t_i-y_i)-(f(x_i)-y_i)\right|
= \frac{1}{m}\sum_{i=1}^m | t_i-f(x_i)| \le\epsilon.
\]
It follows that $T'=\left\{(|t_1-y_1|,\ldots,|t_m-y_m|):t\in T\right\}$
is an $\epsilon$-cover of $L_{\rest{F}{z}}$.
\qed
\fi

\begin{theorem}
Choose a set $X$,
a set $F$ of functions from $X$ to $[0,1]$,
and $\epsilon,\delta>0$.

If there exists $\kappa > 0$ such that for all $\epsilon > 0$,
$\fat_F((1/4-\kappa)\epsilon)$ is finite,
then there is a learner $A$ that $(\epsilon,\delta)$-learns
in the agnostic sense with respect to $F$ from
\begin{equation}
\label{e:fat.agn}
O\left(\frac{1}{\epsilon^2}\left(\fat_F((1/4-\kappa)\epsilon)
        \log^2\frac{1}{\epsilon}+ \log\frac{1}{\delta}\right)\right)
\end{equation}
examples.

If there exists $\kappa > 0$ such that for all $\epsilon > 0$,
$\fatV_F((1/4-\kappa)\epsilon)$ is finite,
then there is a learner $A$ that $(\epsilon,\delta)$-learns
in the agnostic sense with respect to $F$ from
\begin{equation}
\label{e:fatV.agn}
O\left(\frac{1}{\epsilon^2}\left(\frac{1}{\epsilon}
        \fatV_F((1/4-\kappa)\epsilon)\log\frac{1}{\epsilon}
                + \log\frac{1}{\delta}\right)\right)
\end{equation}
examples.
\end{theorem}

{\bf Proof}:
Fix $\beta>0$, a small positive constant.
The algorithm we will consider takes a sample
\[
((x_1,y_1),\ldots,(x_m,y_m))\in(X\times[0,1])^m
\]
and chooses a function $f'\in F$ that has
\[
\frac{1}{m}\sum_{i=1}^m|f'(x_i)-y_i|
<\inf_{f\in F}
\frac{1}{m}\sum_{i=1}^m|f(x_i)-y_i|+\beta.
\]
Fix any distribution $P$ on $X\times[0,1]$.
Let $f^*\in F$ satisfy
$\err_P(f^*) \le \inf_{g\in F} \err_P(g) + \beta$.
% \[
% \int|f^*(x)-y|dP(x,y)\le
% \inf_{g\in F} \int|g(x)-y|dP(x,y) + \beta.
% \]
% to stop the mailer from adding a > in front of it
{From} Hoeffding's inequality,
with probability at least $1-2e^{-2\beta^2m}$ over the sample,
\[
\frac{1}{m}\sum_{i=1}^m|f^*(x_i)-y_i|
\le \int|f^*(x)-y| dP(x,y) + \beta.
\]
If $m\ge\frac{1}{2\beta^2}\log\frac{2}{\delta}$,
this probability is at least $1-\delta/2$.
Applying Theorem~\ref{t:sample} to the class $L_F$ and using
Lemma~\ref{l:losspack},
if
$m$ satisfies (\ref{e:fat.agn}) and (\ref{e:fatV.agn}) above,
every function $f$ in $F$ has
\[
\frac{1}{m}\sum_{i=1}^m|f(x_i)-y_i|-\int|f(x)-y|dP(x,y)
\le\epsilon-3\beta
\]
with probability at least $1-\delta/2$.
% (The proof is identical, except that
% ${\cal N}(\epsilon/2-\alpha,L_{\rest{F}{z}})$ is bounded by
% ${\cal N}(\epsilon/2-\alpha, {\rest{F}{x}})$ using Lemma~\ref{l:losspack}.)
It follows that,
with probability at least $1-\delta$,
\begin{eqnarray*}
\err_P(f')
        & \le & \inf_{f\in F}\frac{1}{m}\sum_{i=1}^m |f(x_i)-y_i|
        +\epsilon-2\beta \\
        & \le & \err_P(f^*) + \epsilon-\beta \\
        & \le & \inf_{g\in F}\err_P(g) + \epsilon.
\end{eqnarray*}
\qed

%6direct.tex

\section{Better bounds in terms of the scale}

In this section, we
describe a more direct approach to bounding the
sample complexity of $\epsilon$-agnostic learning,
which saves a factor of two in the scale at which the dimension
must be finite over that described in the previous section, sometimes 
at the expense of a small increase in the sample complexity.
\begin{theorem}
\label{t:scale}
Choose $X$, a set $F$ of functions from $X$ to $[0,1]$,
and $\epsilon,\delta>0$.

If there is a $\kappa > 0$ such that for all $\epsilon > 0$,
$\fat_F((1/2-\kappa)\epsilon)$ is finite,
then there is a learner $A$ that $(\epsilon,\delta)$-learns in the
agnostic sense with respect to $F$ from
\begin{equation}
\label{e:direct.fat}
O\left(\frac{1}{\epsilon^2}\fat_F((1/2-\kappa)\epsilon)
        \left(\log^2\frac{1}{\epsilon}\right) \left(\log\frac{1}{\delta}\right)\right)
\end{equation}
examples.

If there is a $\kappa > 0$ such that for all $\epsilon > 0$,
$\fatV_F((1/2-\kappa)\epsilon)$ is finite,
then there is a learner $A$ that $(\epsilon,\delta)$-learns in the
agnostic sense with respect to $F$ from
\begin{equation}
\label{e:direct.fatV}
O\left(\frac{1}{\epsilon^3}\fatV_F((1/2-\kappa)\epsilon)
        \left(\log\frac{1}{\epsilon}\right)\left(\log\frac{1}{\delta}\right)\right)
\end{equation}
examples.
\end{theorem}

{\bf Proof}:
Fix $k\in\N$, let $\alpha=1/\lceil 1/(\epsilon \kappa)\rceil$,
and let $\gamma=\alpha/13$.  
Consider a mapping $Q$ from $\left(X\times[0,1]\right)^k\times X^k$
to $[0,1]$, defined as follows.
Fix a function $\phi$ that maps from $X^{2k}$ to
the set of finite subsets of $[0,1]^{2k}$
such that, for any $x\in X^{2k}$,
$\phi(x)$ is a minimal $(\epsilon-9\gamma)$-cover of
$\rest{F}{x}$,
and $\phi(x)$ is invariant under permutations of the components
of $x$.
Then let $x=(x_1,\ldots,x_{2k})\in X^{2k}$, and
for $(y_1,\ldots,y_k)\in[0,1]^k$
let $Q((x_1,y_1),\ldots,(x_k,y_k),x_{k+1},\ldots,x_{2k}) = t'_{2k}$,
where $t'=(t'_1,\ldots,t'_{2k})\in\phi(x)$ satisfies
\[
\frac{1}{k}\sum_{i=1}^k|t'_i-y_i| =
\min_{s\in \phi(x)}\frac{1}{k}\sum_{i=1}^k|s_i-y_i|.
\]

We will first show that, for any distribution on $X\times[0,1]$,
$Q$ predicts the value $y_{2k}$ associated with $x_{2k}$ almost as well
as the best function in $F$, taking expectations over random
sequences.
We use this property to construct a learner that returns a hypothesis
that has error within $\epsilon$ of the best in $F$,
with high probability.

Fix a distribution $P$ on $X\times[0,1]$.
Suppose
\[
((x_1,y_1),\ldots,(x_{2k},y_{2k}))\in \left(X\times[0,1]\right)^{2k}
\]
is a random sequence chosen according to $P$.
Let $x=(x_1,\ldots,x_{2k})$ and
$y=(y_1,\ldots,y_{2k})$.
Choose $f^*\in F$ that satisfies
$\err_P(f^*)\le\inf_{f\in F}\err_P(f) + \gamma$.
% \[
% \int|f^*(x)-y|dP(x,y)\le
% \inf_{f\in F} \int|f(x)-y|dP(x,y) +\gamma.
% \]
In comparing functions defined on $X$,
such as $f^*$,
we will sometimes refer to the function as a vector,
with the obvious interpretation that
$f^*_i=f^*(x_i)$.
Since $\phi(x)$ is an $(\epsilon-9\gamma)$-cover of
$\rest{F}{x}$,
there is a $t^*\in \phi(x)$ with
$\ell_1(t^*,f^*) \le \epsilon-9\gamma$.
It follows that
$\ell_1(t^*,y)\le \epsilon-9\gamma + \ell_1(f^*,y)$.

Applying the Hoeffding bound,
\[
P^{2k}\left\{((x_1,y_1),\ldots,(x_{2k},y_{2k})): 
            \ell_1(f^*,y)> \err_P(f^*)+\gamma\right\} \le 2e^{-4\gamma^2 k}.
\]
If $k>\frac{1}{4\gamma^2}\log\frac{2}{\gamma}$,
this probability is less than $\gamma$.
In that case,
with probability at least $1-\gamma$,
$\ell_1(f^*,y)\le \err_P(f^*)+\gamma$,
% \[
% \ell_1(f^*,y)\le \int|f^*(x)-y|dP(x,y)+\gamma,
% \]
which implies
$\ell_1(t^*,y)\le\epsilon-8\gamma+\err_P(f^*)$.
% \[
% \ell_1(t^*,y)\le\epsilon-8\gamma+\int|f^*(x)-y|dP(x,y).
% \]

For two vectors $a,b\in[0,1]^{2k}$,
define
\[
\ell_1^\first(a,b) = \frac{1}{k}\sum_{i=1}^k |a_i-b_i|,
\]
\[
\ell_1^\last(a,b) = \frac{1}{k}\sum_{i=k+1}^{2k} |a_i-b_i|.
\]
Now, as in the proof of Lemma~\ref{l:twosamp},
let $U$ be the uniform distribution over $\{-1,1\}$.
Then, since $\phi$ is invariant under permutations,
\[
\begin{array}{l}
P^{2k}\left\{((x_1,y_1),\ldots,(x_{2k},y_{2k})): \exists t\in\phi(x)\, 
\left|\ell_1^\first(t,y)-\ell_1^\last(t,y)\right|>2\gamma\right\}\\[3pt]
\hspeq\le \sup_{(x,y)}U^k
\left\{(u_1,\ldots,u_k): \exists t\!\in\!\phi(x)\, 
\left|\frac{1}{k}\sum_{i=1}^k
u_i\left(|t_i\!-\!y_i|\!-\!|t_{i+k}\!-\!y_{i+k}|\right)\right|\!>\!2\gamma\right\}
\end{array}
\]
For any fixed $t\in\phi(x)$,
Hoeffding's inequality implies
\[
U^k\left\{(u_1,\ldots,u_k): 
  \left|\frac{1}{k}\sum_{i=1}^k
   u_i\left(|t_i-y_i|-|t_{i+k}-y_{i+k}|\right)\right|>2\gamma\right\}
   \le 2e^{-2\gamma^2k}.
\]
So with probability at least $1-|\phi(x)|2e^{-2\gamma^2k}$,
for all $t$ in $\phi(x)$,
\begin{equation}        \label{e:goodlyunif}
\left|\ell_1^\first(t,y)-\ell_1^\last(t,y)\right| \leq 2\gamma.
\end{equation}
This implies 
\[
\left|\ell_1^\first(t,y)-\ell_1(t,y)\right| \leq \gamma
\]
and
\[
\left|\ell_1^\last(t,y)-\ell_1(t,y)\right| \leq \gamma.
\]
The probability that this does not happen is no more than $\gamma$ if
\[
\sup_{x\in X^{2k}}
{\cal N}\left(\epsilon-9\gamma,\rest{F}{x}\right)
2e^{-2\gamma^2 k}\le\gamma.
\]
Now, Lemmas~\ref{l:quantok} and~\ref{l:fat.packbound}, together
with (\ref{e:wellknown}), imply that
\begin{eqnarray*}
{\cal N}\left(\epsilon-9\gamma,\rest{F}{x}\right)
        & \le & {\cal N}\left(\epsilon-10\gamma,
                            Q_\gamma\left(\rest{F}{x}\right)
                \right) \\
% 2 * 13 * e * 2 < 142
  & \le & 2\left(\frac{1}{\gamma}\right)^{6d\log_2(142 k/(d\gamma))},
\end{eqnarray*}
% %
% %provided that $2k\ge d = \fat_F(\epsilon-13\gamma)$,
% provided that
% % 4 * 13 * e / ln 2 < 204
% \begin{equation}
% \label{e:provided}
% 2k\ge d \log_2 (204/\gamma),
% \end{equation}
where $d = \fat_F(\epsilon-13\gamma)$,
since $\fat_{Q_\gamma(F)}(\epsilon-12\gamma) \le \fat_F(\epsilon-13\gamma)$.
So Inequality~(\ref{e:goodlyunif}) will hold for all $t$
in $\phi(x)$ with probability at least $1-\gamma$,
provided
\[
4\exp\left(\frac{6d}{\ln 2}\left(\ln\frac{142 k}{d\gamma}\right)
                           \left(\ln\frac{1}{\gamma}\right)
- 2\gamma^2 k\right)\le \gamma.
\]
Applying Lemma~\ref{l:once.and.for.all.again}, 
we can see that there is a constant $c$ such that
\[
k\ge\frac{c}{\gamma^2}\left(d \ln^2 \frac{1}{\gamma}
+\ln\frac{1}{\gamma}\right)
\]
will suffice.
In that case,
with probability at least $1-2\gamma$, the $t'\in\phi(x)$ with
minimal $\ell_1^\first(t',y)$
satisfies
\begin{eqnarray*}
\ell_1(t',y)
        & \le & \ell_1^\first(t',y)+\gamma\\
        & \le & \ell_1^\first(t^*,y)+\gamma\\
        & \le & \ell_1(t^*,y) + 2\gamma\\
        & \le & \epsilon-6\gamma + \err_P(f^*),
\end{eqnarray*}
and hence
\begin{eqnarray*}
\ell_1^\last(t',y)
        & \le & \epsilon-5\gamma+ \err_P(f^*)\\
        & \le & \epsilon-4\gamma+ \inf_{f\in F}\err_P(f).
\end{eqnarray*}
That is,
\[
P^{2k}\big\{
((x_1,y_1),\ldots,(x_{2k},y_{2k})):
\ell_1^\last(t',y) > \epsilon-4\gamma
   + \inf_{f\in F}\err_P(f) \big\} < 2 \gamma,
\]
which implies
\[
\int\ell_1^\last(t',y)dP^{2k}((x_1,y_1),\ldots,(x_{2k},y_{2k})) 
 < \epsilon-2\gamma+\inf_{f\in F}\err_P(f),
\]
and hence
\[
\int(|Q((x_1,y_1),\ldots,(x_k,y_k),x_{k+1},\ldots,x_{2k})-y_{2k}| 
 dP^{2k}((x_1,y_1),\ldots,(x_{2k},y_{2k}))
- \inf_{f\in F} \err_P(f)
 < \epsilon-2\gamma.
\]
If we define the hypothesis $h$ of $Q$ as
\[
h(\beta) = Q((x_1,y_1),\ldots,(x_k,y_k),x_{k+1},\ldots,x_{2k-1},\beta),
\]
we have
\[
P^{2k-1}\left\{
((x_1,y_1),\ldots,(x_{2k-1},y_{2k-1})): 
 \err_P(h)-\inf_{f\in F}\err_P(f) > \epsilon-\gamma \right\} 
     < 1-\gamma/\epsilon.
\]

To complete the proof, we use a technique
from~\cite{hklw} to convert this prediction strategy
to an agnostic learning algorithm.
Consider the algorithm which takes as input
$N_1(2k-1)+N_2$ labelled examples,
uses the first $N_1(2k-1)$ examples and the mapping $Q$ to
compute $N_1$ hypotheses,
and outputs the hypothesis from this set that has minimum error
over the remaining $N_2$ examples.
With probability at least $1-\delta/2$,
at least one of the $N_1$ hypotheses has error no more than
$\epsilon-\gamma$,
provided that $(1-\gamma/\epsilon)^{N_1} < \delta/2$;
setting
$N_1=\frac{\epsilon}{\gamma}\ln\frac{2}{\delta}$
will suffice for this.
For each of these $N_1$ hypotheses $h$, the algorithm calculates the
empirical error
% \[
$\frac{1}{N_2}\sum_{i=1}^{N_2}|h(x_i)-y_i|$,
% \]
and chooses the hypothesis with the minimum empirical error.
Hoeffding's inequality implies that the probability
that some hypothesis has empirical error more than $\gamma/2$
from $\err_P(h)$ is no more than $2N_1e^{-\gamma^2N_2/2}$.
This probability is less than $\delta/2$ when 
% \[
$N_2>\frac{2}{\gamma^2}\log\frac{4N_1}{\delta}$.
% \]
This implies that, with probability at least $1-\delta$ over the
$N_1(2k-1)+N_2$ examples,
the hypothesis returned by the algorithm has error less than
$\epsilon$.
Clearly,
the algorithm needs to see
\[
O\left(\frac{\epsilon d}{\alpha^3}\log\frac{1}{\delta}
        \log^2\frac{1}{\alpha} +
        \frac{1}{\alpha^2}\log\frac{\epsilon}{\delta\alpha}\right)
\]
examples, completing the proof of (\ref{e:direct.fat}).
The bound (\ref{e:direct.fatV}) can be proved analogously
using Lemma~\ref{l:fatv.packbound} in place of Lemma~\ref{l:fat.packbound}.
\qed

Buescher and Kumar proposed a related algorithm in~\cite{bk92}.
Their algorithm (the ``canonical estimator'') splits a sequence of
labelled examples into two parts.
Let $\xi$ be the sequence of points from $X$ in the first part of the
sample.
The algorithm chooses a finite subset
% \footnote{The requirement that $T$ is a subset of $F$,
% although not explicit in their description of the canonical estimator,
% is clearly necessary}
$T$ of $F$ such that
$\rest{T}{\xi}$ is a cover of $\rest{F}{\xi}$.
It then returns the function in $T$ that has minimal error on the
remaining part of the sample.
Interestingly, this algorithm also discards the labels of part of the
training sample.

%PL - I think referee is right, algorithm may be improper
%
% A proof of (\ref{e:wellknown}), which bounds covering numbers in terms
% of packing numbers, constructs a covering using elements of the packing.
% If the covering used by the algorithm in the proof of Theorem~\ref{t:scale}
% is constructed in such a manner, one can prove the bounds of that
% theorem using an algorithm which only hypothesizes functions in $F$.

% 7necc.tex

\section{Necessary conditions}

In this section, we collect necessary conditions for some
of the properties considered in this paper.  Coupled with
the positive results of the previous sections, these
results considerably narrow the constant factor gap
between the scales at which the finiteness of the
scale-sensitive dimensions is necessary and sufficient
for learning and
% GCness
the GC property.
We also provide examples showing
that these necessary conditions are not sufficient conditions,
and that they cannot be improved.

First, we prove the necessity condition for $\epsilon$-uniform
GC classes.  The proof is based on that of the analogous
result for $\fatV$ which was proved in \cite{abch} and
follows from this new result since $\fatV_F \leq \fat_F$
for all $F$.
It improves on the result in~\cite{abch}
by a factor of $2$ the scale
at which $\fat_F$'s finiteness is necessary for
$F$ to be an $\epsilon$-uniform GC class.
\begin{theorem}
\label{t:gc.necc}
Choose $X$, $F \subseteq [0,1]^X$, and $0 < \epsilon < 1$.  Then if
there exists $\alpha > 0$ such that $\fat_F(\epsilon/2 + \alpha) = \infty$,
then $F$ is not an $\epsilon$-uniform GC class.
\end{theorem}
{\bf Proof}:  Choose $0 < \epsilon < 1$.
Assume for contradiction that there exist $X$, $F \subseteq [0,1]^X$, and
$\alpha > 0$ such that $\fat_F (\epsilon/2 + \alpha) = \infty$ but
that $F$ is an $\epsilon$-uniform GC class.  Let
$m = m_{\mathrm{GC},F}(\epsilon,1/2)$.  Choose $d \in \N$ such that
\begin{equation}
\label{e:dbig}
d \geq \frac{m}{\alpha} (1 + \epsilon/2 + \alpha).  
\end{equation}
Let $(x_1,r_1),...,(x_d,r_d)$ be $(\epsilon/2+\alpha)$-fatly shattered
by $F$, and let $D$ be the uniform distribution over
$x_1,...,x_d$.  Let $r = (1/d) \sum_{i=1}^d r_i$.

We claim that for {\em any} sequence $u_1,...,u_m$ of elements
of $\{ x_1,...,x_d \}$, there is an $f \in F$ such that
\[
\left| \left( \frac{1}{m} \sum_{i=1}^m f(u_m) \right)
        - \int f(z)\; dD(z) \right| > \epsilon.
\]
Choose such a $u_1,...,u_m$, and for each $j \in \{ 1,...,m \}$
let $i_j$ be such that $u_j = x_{i_j}$.

Assume as a first case that
\begin{equation}
\label{e:gc.necc.1}
\frac{1}{m} \sum_{j=1}^m r_{i_j} \leq r.
\end{equation}
Choose $f \in F$ such that
\[
f(x_i) \left\{ \begin{array}{ll}
                  \leq r_i - (\epsilon/2 + \alpha) 
                      & \mbox{if $i \in \{ i_1,...,i_m \}$} \\
                  \geq r_i + (\epsilon/2 + \alpha) 
                      & \mbox{otherwise} 
              \end{array}
        \right.
\]
Then
\begin{eqnarray*}
\frac{1}{m} \sum_{j=1}^m f(x_{i_j})
      & \leq & \frac{1}{m} \sum_{j=1}^m (r_{i_j} - (\epsilon/2 + \alpha)) \\
      & \leq & r - (\epsilon/2 + \alpha)
\end{eqnarray*}
by (\ref{e:gc.necc.1}).  However,
\begin{eqnarray*}
\int f(z) \; dD(z)
        & = & \frac{1}{d} \sum_{i=1}^d f(x_i) \\
        & \geq & \frac{1}{d} \sum_{i \not\in \{ i_1,...,i_m \}} f(x_i) \\
        & \geq & \frac{1}{d} \sum_{i \not\in \{ i_1,...,i_m \}} r_i
                                        + (\epsilon/2 + \alpha) \\
        & = & \frac{d-m}{d}\left( \frac{1}{d-m}
                               \sum_{i \not\in \{ i_1,...,i_m \}} r_i
                                        + (\epsilon/2 + \alpha) \right) \\
        & \geq & \frac{d-m}{d}\left( r+ (\epsilon/2 + \alpha) \right) 
\end{eqnarray*}
by (\ref{e:gc.necc.1}) together with the definition of $r$.  Thus,
\begin{eqnarray*}
\int f(z) \; dD(z) - \frac{1}{m} \sum_{j=1}^m f(u_j)
        & \geq & (1-m/d)(r+\epsilon/2+\alpha) - (r-(\epsilon/2+\alpha))\\
        & = & \epsilon+ 2\alpha-\frac{m}{d}(r+\epsilon/2+\alpha)\\
        & \geq & \epsilon+\alpha,
\end{eqnarray*}
from (\ref{e:dbig}), 
completing the proof in this case.  
The case in which $\frac{1}{m} \sum_{j=1}^m r_{i_j} > r$ can be
handled similarly.

Therefore, we have that for samples of size $m$, there is a function
in $F$ whose expectation is estimated with accuracy worse than $\epsilon$,
a contradiction, completing the proof. \qed

The next result shows that this condition is not sufficient 
for $F$ to be an $\epsilon$-uniform GC class.

\begin{theorem}
\label{t:fat.GC.notsuff}
For each $0<\epsilon<1/2$,
there is a function class $F$ that is not an $\epsilon$-uniform GC
class, but for all $\alpha>0$, $\fat_F(\epsilon/2+\alpha)$ is finite.
\end{theorem}

{\bf Proof}:
Fix $0<\epsilon<1/2$ and let $F$ be the class of all functions $f$
from $\N$ to $[0,1]$ satisfying
\[
f(i) \in\{1/2 + (\epsilon/2 + 1/(i+3)),1/2 - (\epsilon/2 + 1/(i+3))\}.
\]
Clearly, for all $\alpha>0$, $\fat_F(\epsilon/2+\alpha)$ is finite.
For sample size $m$,
consider the distribution $D$ that is uniform on
%PL
% $Z=\{1,\ldots,m^2 e^{\epsilon m+2}\}$.
$Z=\{1,\ldots,m^2 e^{\epsilon m}\}$.
Then for any sequence $x_1,\ldots,x_m$,
there is a function $f$ in $F$ with
\[
\left|\frac{1}{m}\sum_{i=1}^m f(x_i) - \int_X f(x) dD(x) \right| > \epsilon.
\]
To see this, fix a sequence
$x_1,\ldots,x_m$,
let $d=(m+3)^2 e^{\epsilon m}$,
and consider the function $f$ in $F$ satisfying
\[
f(n) = \left\{\begin{array}{ll}
        1/2 - (\epsilon/2 + 1/(n+3)) & \mbox{if $n=x_i$ for some $i$}\\
        1/2 + \epsilon/2 + 1/(n+3) & \mbox{otherwise.}
\end{array}\right.
\]
If we define $S_x=\{x_1,\ldots,x_m\}$, then
\[
\begin{array}{l}
%PL switched these two - as referee pointed out
\int_X f(x) dD(x) - \frac{1}{m}\sum_{i=1}^m f(x_i) \\[3pt]
\hspeq> \frac{1}{d}\left(\sum_{i=1}^d f(i)\right) - 1/2 +
                                \epsilon/2\\[3pt]
\hspeq= \frac{1}{d}\left(\sum_{i=1}^d f(i) - 1/2 +
                                \epsilon/2 \right) \\[3pt]
\hspeq= \frac{1}{d}\left(\sum_{i\in S_x}
                                (f(i)-1/2+\epsilon/2) 
                            +\sum_{i\in Z- S_x}
                                (f(i)-1/2+\epsilon/2)\right)\\[3pt]
\hspeq= \frac{1}{d}\left(\sum_{i\in S_x}
                                (-1/(i+3)) + \sum_{i\in Z- S_x} (\epsilon+1/(i+3))\right).
\end{array}
\]
Both sums are clearly minimized when $S_x=\{1,\ldots,m\}$.
Using the fact that $\ln(n+1)<\sum_{i=1}^n 1/i<\ln n + 1$,
we have
\[
%PL switched
\int_X f(x) dD(x) - \frac{1}{m}\sum_{i=1}^m f(x_i) > \epsilon +
\left(\ln(d+4) - \epsilon m -2\ln (m+3)+4\right)/d,
\]
but the definition of $d$ implies that this quantity is at least
$\epsilon$.
\qed

Note that for all function classes $F$ and $\epsilon>0$,
$\fat_F(\epsilon) \geq \fatV_F(\epsilon)$, so that
Theorem~\ref{t:gc.necc}  implies the same thing about $\fatV$.
The following observation shows that
there is no better necessary condition in terms of $\fat$
or $\fatV$.

\begin{proposition}
\label{p:fatV.GC.notnecc}
There is a function class $F$ that is an $\epsilon$-uniform GC class,
but has $\fatV_F(\epsilon/2)$ infinite.
\end{proposition}

{\bf Proof}:
Suppose $F$ is the set of all functions from the natural numbers to
$\{ 1/2 - \epsilon/2, 1/2 + \epsilon/2 \}$.  Clearly, 
$\fatV_F(\epsilon/2) = \infty$.  However, for any sample, the
estimate of the expectation of any member $f$ of $F$ must be in
$[1/2 - \epsilon/2, 1/2 + \epsilon/2]$, as must be the true expectation
of $f$ with respect to any distribution.
\qed

Next, we turn to proving a necessary condition for $\epsilon$-agnostic
learnability.  The following variant of $\fat_F$, due to
Simon~\cite{simon},
will be useful.  For $X$, $F \subseteq [0,1]^X$, and $\gamma > 0$, 
we say $F$ {\em strongly $\gamma$-fatly shatters}
a sequence $(x_1,l_1,u_1),...,(x_d,l_d,u_d)$
of elements of $X \times [0,1]^2$
if $u_i\ge l_i+2\gamma$ for $i=1,\ldots,d$ and,
for all $(b_1,...,b_d) \in \{ 0, 1 \}^d$, there is an
$f \in F$ such that
\[
\begin{array}{l}
f(x_i) = u_i \Leftrightarrow b_j = 1 \\
f(x_i) = l_i \Leftrightarrow b_j = 0
\end{array}
\]
for $i=1,\ldots,d$.
We then define $\sfat_F (\gamma)$ to be the length of the longest
sequence that is strongly $\gamma$-fatly shattered by $F$,
or $\infty$ is there is no longest sequence.

The following lemma,
whose proof closely follows parts of that of a related
result in~\cite{simon}, as well as Theorem~\ref{t:binfun.low},
will be useful.
\begin{lemma}
\label{l:sfat.necc}
Choose $X$, $F \subseteq [0,1]^X$, $\epsilon > 0$.  Then if there
exists $\alpha > 0$ such that $\sfat_F (\epsilon + \alpha)$
is infinite, then $F$ is not $\epsilon$-agnostically learnable.
\end{lemma}
{\bf Proof}:
Assume for contradiction that $F$ is 
$\epsilon$-agnostically learnable, but that there
exists $\alpha > 0$ such that $\sfat_F (\epsilon + \alpha)$
is infinite.  Fix such an $\alpha > 0$.  
Let $m \in \N$, and a learner $A$ be such that for all distributions
$P$ on $X \times [0,1]$,
\[
P^m \left\{ z : \int |(A(z))(x) - y|\; dP(x,y) 
  \geq (\inf_{f \in F} |f(x)-y|)+\epsilon \right\} \leq 1/2.
\]

Choose $d \in \N$ such that
\begin{equation}
\label{e:sfat.dbig}
d > \frac{m (\epsilon + \alpha)}{\alpha}.
\end{equation}
Choose a sequence $(x_1,l_1,u_1),\ldots,(x_d,l_d,u_d)$ from among
those $(\epsilon + \alpha)$-fatly shattered by $F$.  
For each $b \in \{ 0, 1\}^d$, choose $f_b \in F$ so that
\[
\begin{array}{l}
f_b(x_i) = u_i \Leftrightarrow b_i = 1 \\
f_b(x_i) = l_i \Leftrightarrow b_i = 0,
\end{array}
\]
and let $G = \left\{ f_b : b \in \{ 0, 1 \}^d\right\}$.
For each $b \in \{ 0, 1\}^d$,
let $P_b$ be a distribution over $X \times [0,1]$
obtained by choosing the first component uniformly from $x_1,...,x_d$,
and evaluating $f_b$ at the first component to get the second.
Note that for each such $P_b$, $\inf_{f \in F} \int |f(x)-y|\;dP_b(x,y) = 0$.

% Assume wlog are distinct
Choose $v_1,...,v_m \in \{ x_1,...,x_d \}$,
and let $i_1,...,i_m$ be such that for each $1 \leq j \leq m$,
$v_j = x_{i_j}$.
Notice that 
$h_b = A((v_1,f_b(v_1)),...,(v_m,f_b(v_m)))$
is a function only of those components $b_i$ for
which $i \in \{ i_1,...,i_m \}$.
Suppose we choose $b$ uniformly
according to the uniform distribution over $\{ 0, 1\}^d$.
Choose $i \not \in \{ i_1,...,i_m \}$,
and $(c_1,...,c_m) \in \{ -1, 1\}^m$.
Then, by the independence of the choice of
$b_i$ from that of the other components, in particular those determining $h_b$,
the expectation of $|h_b (x_i) - f_b (x_i)|$, 
given that $b_{i_1} = c_1,...,b_{i_m} = c_m$, is
\[
1/2 | h_b (x_i) - u_i| + 1/2 | h_b (x_i) - l_i|\ge
1/2 | u_i-l_i|\ge\epsilon+\alpha.
\]
Since this is true independent of $c_1,...,c_m$, for any
$i \not\in \{ i_1,...,i_m \}$, the expected value of the
error of $A$ on $x_i$ is at least $\epsilon + \alpha$.  Therefore,
the overall expected error of $A$'s hypothesis, over the random choice
of $b$, is at least 
\[
(1-m/d) (\epsilon + \alpha).
\]
This implies there exists $b$ such that 
if 
\[
z = ((v_1,f_b(v_1)),...,(v_m,f_b(v_m))),
\]
$\int |(A(z))(x)-y| P_b (x,y)$ is at least
\[
(1-m/d) (\epsilon + \alpha).
\]
Since $v_1,...,v_m$ was chosen arbitrarily, by (\ref{e:sfat.dbig}),
this contradicts the fact that $A$ $(\epsilon,1/2)$-agnostically
learns, completing the proof.
\qed

\begin{theorem}
\label{t:fat.agn.necc}
Choose $X$, $F \subseteq [0,1]^X$, and $\epsilon>0$.  Then if
there exists $\alpha > 0$ such that $\fat_F(\epsilon + \alpha)$ is
infinite, then $F$ is not $\epsilon$-agnostically learnable.
\end{theorem}

{\bf Proof}:
Fix $\alpha>0$ such that $\fat_F(\epsilon+\alpha)$ is infinite.
Then $\fat_{Q_{\alpha/3}(F)}(\epsilon+2\alpha/3)$ is infinite.
By Lemma~9 of~\cite{ab-euro}, this implies
$\sfat_{Q_{\alpha/3}(F)}(\epsilon+2\alpha/3)$ is infinite,
and then Lemma~\ref{l:sfat.necc} implies $Q_{\alpha/3}(F)$ is not
$(\epsilon+\alpha/3)$-agnostically learnable.
But then $F$ is not $\epsilon$-agnostically learnable,
since for every $f\in F$ and distribution $P$ on $X\times[0,1]$,
$\err_P(f)\le\err_P(Q_{\alpha/3}(f))+\alpha/3$,
so a learner that $\epsilon$-agnostically learns $F$ can
$(\epsilon+\alpha/3)$-agnostically learn $Q_{\alpha/3}(F)$.
\qed

Next, we show that the converse of Theorem~\ref{t:fat.agn.necc} 
is not true.
% The proof, which uses elements of the proofs
% of Theorems~\ref{t:binfun.low} and \ref{t:fat.GC.notsuff}, and the work of 
% Simon \cite{simon}, is omitted.
\begin{theorem}
\label{t:fat.agn.notsuff}
For each $0<\epsilon<1/4$,
there is a function class $F$ that is not $\epsilon$-agnostically
learnable,
but for all $\alpha>0$, $\fat_F(\epsilon+\alpha)$ is finite.
\end{theorem}
{\bf Proof}:
As in the proof of Theorem~\ref{t:fat.GC.notsuff},
fix $0<\epsilon<1/4$ and let $F$ be the class of all functions $f$
from $\N$ to $[0,1]$ satisfying
\[
f(i) \in\{1/2 + (\epsilon + 1/(i+3)),1/2 - (\epsilon + 1/(i+3))\}.
\]
Clearly, for all $\alpha>0$, $\fat_F(\epsilon+\alpha)$ is finite.

Choose $d \in \N$.  For $b \in \{ -1, 1 \}^d$, 
choose $f_b$ such that for each $i \in \{ 1,..,d \}$,
$f(i) = 1/2 + b_i (\epsilon + 1/(i+3))$.  Define a distribution
$P_b$ over $\{ 1,...,d\} \times [0,1]$ by choosing the
first component uniformly from $\{ 1,...,d \}$, and evaluating
$f_b$ at the first component to get the second.

Arguing as in Lemma~\ref{l:sfat.necc}, we can see that for any
algorithm $A$ and any $x_1,...,x_m$, 
if for all $b$, 
\[
h_b = A((x_1,f_b(x_1)),...,(x_m,f_b(x_m))),
\]
if $i$ is not in the sample and $b$ is chosen uniformly at
random, then the expectation of $A$'s error is at least
$\epsilon+1/(i+3)$.

Arguing as in Theorem~\ref{t:fat.GC.notsuff}, if $d$ is large
enough, this expected error is greater than $\epsilon$, whatever
the value of $x_1,...,x_m$.  Therefore, there exists a $b$ for which
this is true, completing the proof.
\qed

Next, we observe that none of Theorem~\ref{t:fat.agn.necc} and
its corollaries with regard to $\fatV$ or agnostic learning
can be improved.

\begin{proposition}
\label{p:fatV.agn.notnecc}
There is a function class $F$ that is $\epsilon$-agnostically
learnable, but has $\fatV_F(\epsilon)$ infinite.
\end{proposition}

{\bf Proof}:
Suppose $F$ is the set of all functions from the natural numbers to
$\{ 1/2 - \epsilon, 1/2, 1/2 + \epsilon \}$.  Clearly, 
$\fatV_F(\epsilon) = \infty$.  However, the hypothesis of
a constant $1/2$ is always $\epsilon$-close to any $f \in F$, and
therefore an algorithm that simply outputs this hypothesis
$\epsilon$-agnostically learns $F$.
\qed

\begin{figure}
\begin{center}
\includegraphics{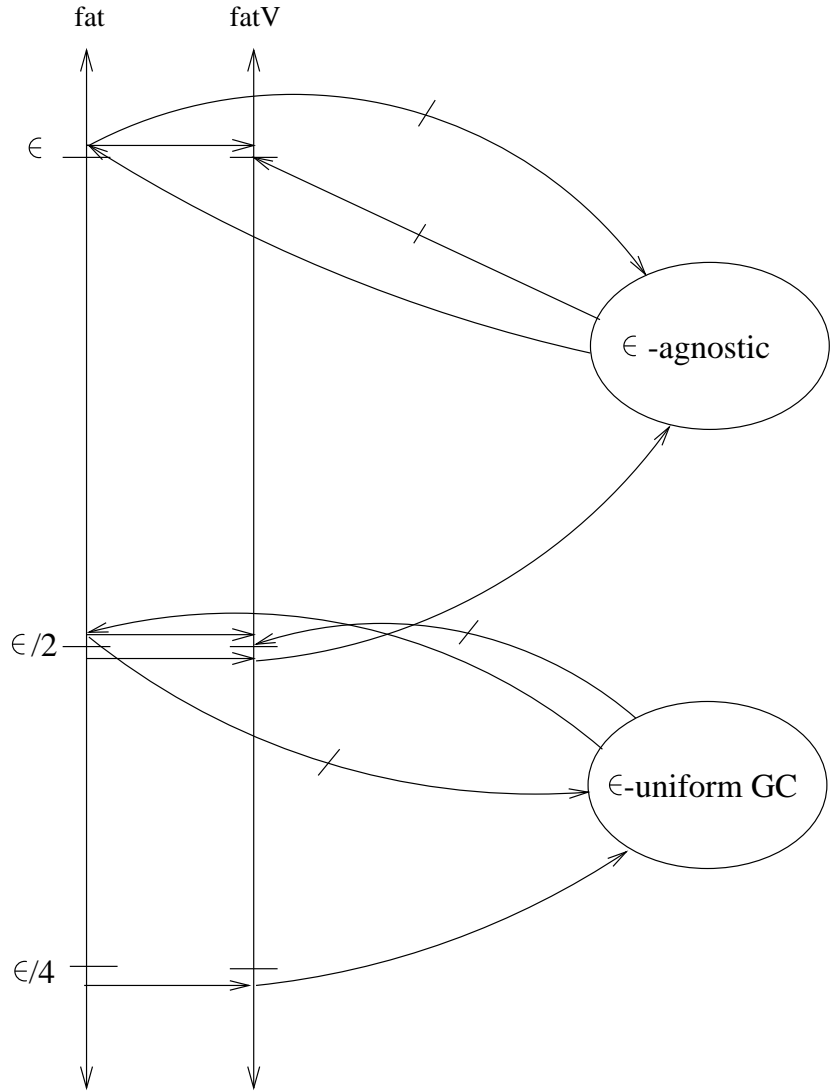}
\end{center}
\caption{This figure represents the state of our knowledge with
  regard to the relationship between the finiteness of $\fat$ and $\fatV$
  at certain scales and learnability and uniform convergence.  A point on
  one of the number lines corresponding to $\fat$ or $\fatV$ at position
  $\gamma$ on the line represents the statement ``$\fat_F(\gamma)$
  (respectively $\fatV_F(\gamma)$) is finite''.  The ellipses on the right
  have the obvious interpretation.  An arrow indicates an implication, a
  crossed out arrow indicates that no such implication exists.}
\label{f:results}
\end{figure}

Our results about the relationship between the finiteness
of $\fatV$ and $\fat$, and
% $\epsilon$-uniform GCness,
the $\epsilon$-uniform GC property and $\epsilon$-agnostic
learnability, are summarized in Figure~\ref{f:results}.

%  \begin{figure}
%  \begin{center}
%  \mbox{\psfig{figure=lotsa.arrows.eps,height=4.73in,width=3.07in}}\\
%  \end{center}
%  \caption{This figure represents the state of our knowledge with
%  regard to the relationship between the finiteness of $\fat$ and
%  $\fatV$ at certain scales and learnability and uniform convergence.
%  A point on one of the number lines corresponding to $\fat$ or $\fatV$
%  at position $\gamma$ on the line represents the statement 
%  ``$\fat_F(\gamma)$ is finite''.  The ellipses on the right have
%  the obvious interpretation.  An arrow indicates that we know of
%  an implication, a crossed out arrow indicates that we know no
%  such implication exists.}
%  \label{f:results}
%  \end{figure}

% 99end.tex
% Acknowledgements, references, etc

\section*{Acknowledgements}

We thank Pankaj Agarwal, David Haussler, Wee Sun Lee, and T.M. Murali
for their help, and two anonymous referees for their comments.  Peter
Bartlett was supported by the Australian Telecommunications and
Electronics Research Board.  This work was done while Phil Long was at
Duke University supported by US Office of Naval Research grant
N00014--94--1--0938 and US Air Force Office of Scientific Research
grant F49620--92--J0515.

%end original article

\bibliographystyle{plain}
\bibliography{bib}

\end{document}